\documentclass[journal,11pt,onecolumn,draftclsnofoot,]{IEEEtran}

\usepackage[letterpaper, left=1.25in, right=1.25in, bottom=1in, top=1in]{geometry}

\usepackage{macros}

\usepackage{capt-of}

\usepackage{multirow}

\usepackage{hyperref}
\usepackage[capitalise, nameinlink]{cleveref}
\hypersetup{bookmarksopen=true}

\crefformat{equation}{\textup{#2(#1)#3}}
\crefrangeformat{equation}{\textup{#3(#1)#4--#5(#2)#6}}
\crefmultiformat{equation}{\textup{#2(#1)#3}}{ and \textup{#2(#1)#3}}
{, \textup{#2(#1)#3}}{, and \textup{#2(#1)#3}}
\crefrangemultiformat{equation}{\textup{#3(#1)#4--#5(#2)#6}}%
{ and \textup{#3(#1)#4--#5(#2)#6}}{, \textup{#3(#1)#4--#5(#2)#6}}{, and \textup{#3(#1)#4--#5(#2)#6}}

\Crefformat{equation}{#2Equation~\textup{(#1)}#3}
\Crefrangeformat{equation}{Equations~\textup{#3(#1)#4--#5(#2)#6}}
\Crefmultiformat{equation}{Equations~\textup{#2(#1)#3}}{ and \textup{#2(#1)#3}}
{, \textup{#2(#1)#3}}{, and \textup{#2(#1)#3}}
\Crefrangemultiformat{equation}{Equations~\textup{#3(#1)#4--#5(#2)#6}}%
{ and \textup{#3(#1)#4--#5(#2)#6}}{, \textup{#3(#1)#4--#5(#2)#6}}{, and \textup{#3(#1)#4--#5(#2)#6}}

\crefname{fq}{Fundamental Question}{Fundamental Questions}

\usepackage[inline]{enumitem}

\usepackage[most]{tcolorbox}
\usepackage{multicol}

\let\[\relax \let\]\relax 
\DeclareRobustCommand{\[}{\begin{equation}}
\DeclareRobustCommand{\]}{\end{equation}}

\begin{document}

\bstctlcite{IEEEexample:BSTcontrol}

\title{Deep Learning Meets Sparse Regularization: \\ A Signal Processing Perspective}

\author{Rahul~Parhi,~\IEEEmembership{Member,~IEEE},
and~Robert~D.~Nowak,~\IEEEmembership{Fellow,~IEEE}%
\thanks{Rahul Parhi is with the Biomedical Imaging Group, \'Ecole Polytechnique F\'ed\'erale de Lausanne, Lausanne, Switzerland (e-mail: rahul.parhi@epfl.ch).

Robert D. Nowak is with the Department of Electrical and Computer Engineering, University of Wisconsin--Madison,  Madison, WI, USA (e-mail: rdnowak@wisc.edu).}}

\markboth{}{}

\maketitle

\begin{abstract}
    Deep learning has been wildly successful in practice and most state-of-the-art machine learning methods are based on neural networks. Lacking, however, is a rigorous mathematical theory that adequately explains the amazing performance of deep neural networks. In this article, we present a relatively new mathematical framework that provides the beginning of a deeper understanding of deep learning. This framework precisely characterizes the functional properties of neural networks that are trained to fit to data. The key mathematical tools which support this framework include transform-domain sparse regularization, the Radon transform of computed tomography, and approximation theory, which are all techniques deeply rooted in signal processing. This framework explains the effect of weight decay regularization in neural network training, the use of skip connections and low-rank weight matrices in network architectures, the role of sparsity in neural networks, and explains why neural networks can perform well in high-dimensional problems. 

\end{abstract}

\IEEEpeerreviewmaketitle

\section{Introduction}
\IEEEPARstart{D}{eep} learning (DL) has revolutionized engineering and the sciences in the modern, data age. The typical goal of DL is to predict an output $y \in \Y$ (e.g., a label or response) from an input $\vec{x} \in \X$ (e.g., a feature or example). A neural network (NN) is ``trained'' to fit to a set of data consisting of the pairs $\{(\vec{x}_n,y_n)\}_{n=1}^N$, by finding a set of NN parameters $\vec{\theta}$ so that the NN mapping closely matches the data. The trained NN is a function, denoted by $f_\vec{\theta}: \X \to \Y$, that can be used to predict the output $y \in \Y$ of a new input $\vec{x} \in \X$. This paradigm is referred to as \emph{supervised learning}, which is the focus of this article. The success of deep learning has spawned a burgeoning industry that is continually developing new applications, NN architectures, and training algorithms. 
This article reviews recent developments in the mathematics of DL, focused on the characterization of the kinds of functions learned by NNs fit to data. There are currently many competing theories to explain the success of DL. These developments are part of a wider body of theoretical work that can be crudely organized into three broad categories:
\begin{enumerate*}
    \item approximation theory with NNs;
    \item the design and analysis of optimization (``training'') algorithms for NNs; and
    \item characterizations of the properties of trained NNs.
\end{enumerate*}

This article belongs to the latter category of research and investigates the functional properties (i.e., the \emph{regularity}) of solutions to NN training problems with \emph{explicit}, Tikhonov-type regularization.  
While much of the success of deep learning in practice comes from networks with highly structured architectures, it is hard to establish a rigorous and unified theory for such NNs used in practice. 
Therefore, we primarily focus on fully-connected, feedforward NNs with the popular Rectified Linear Unit (ReLU) activation function. This article introduces a mathematical framework that unifies a line of work from several authors over the last few years that sheds light on the nature and behavior of NN functions which are trained to a global minimizer with explicit regularization. The presented results are just one piece of the puzzle towards developing a mathematical theory of deep learning.
The purpose of this article is, in particular, to provide a gentle introduction to this new mathematical framework, accessible to readers with a mathematical background in \emph{Signals and Systems} and \emph{Applied Linear Algebra}. The framework is based on mathematical tools familiar to the signal processing community, including transform-domain sparse regularization, the Radon transform of computed tomography, and approximation theory. It is also related to well-known signal processing ideas such as wavelets, splines, and compressed sensing. This framework provides a new take on the following fundamental questions:
\begin{enumerate}
    \item What is the effect of regularization in DL?
    \item What kinds of functions do NNs learn?
    \item What is the role of NN activation functions?
    \item Why do NNs seemingly break the curse of dimensionality?
\end{enumerate}

\section{Neural Networks and Learning from Data}
\begin{figure}[t]
    \centering
    \begin{minipage}[b]{0.3\linewidth}
        \centering
        \centerline{\includegraphics[width=\textwidth]{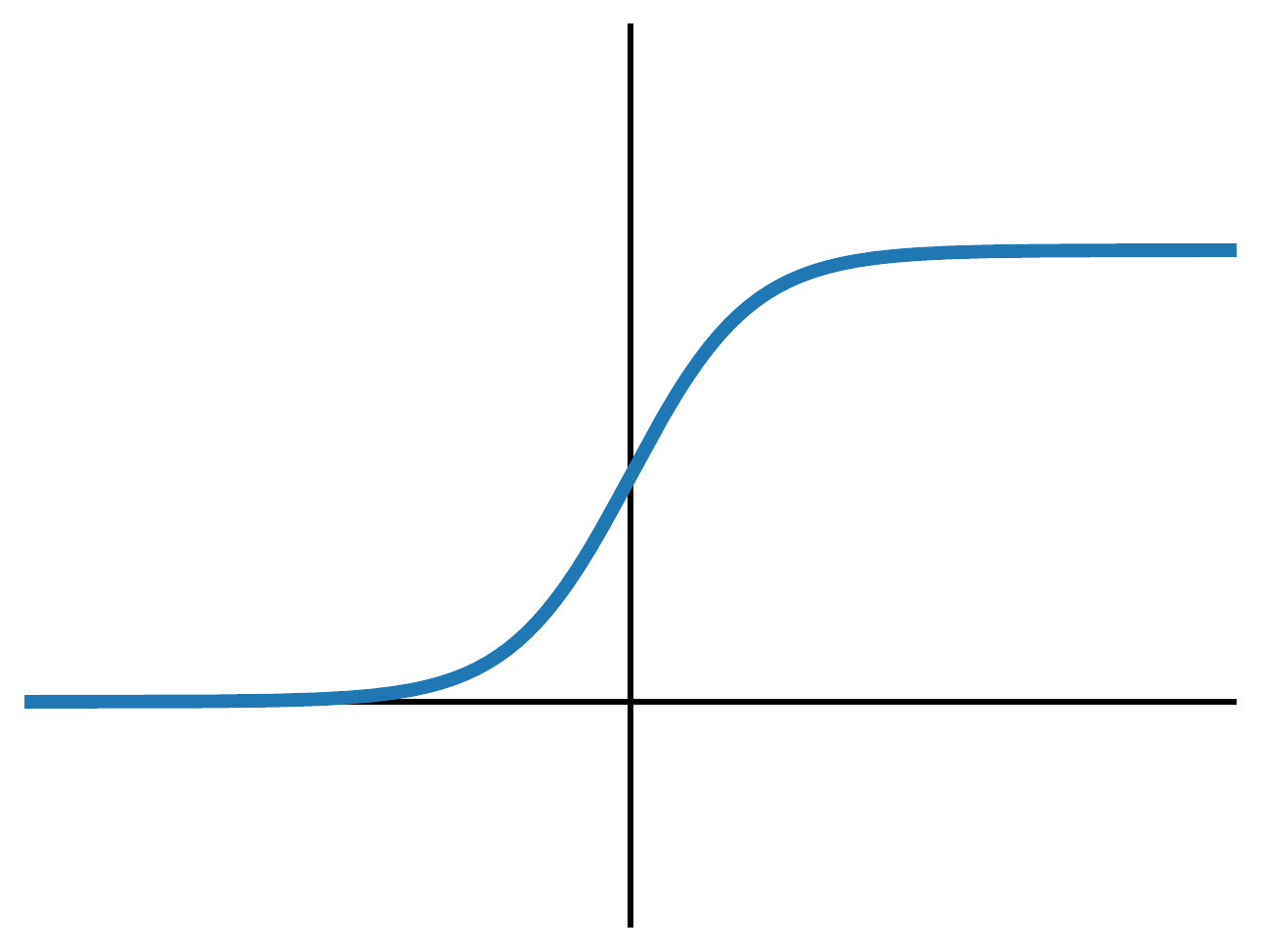}}
        (a) Sigmoid
    \end{minipage}
    \begin{minipage}[b]{0.3\linewidth}
        \centering
        \centerline{\includegraphics[width=\textwidth]{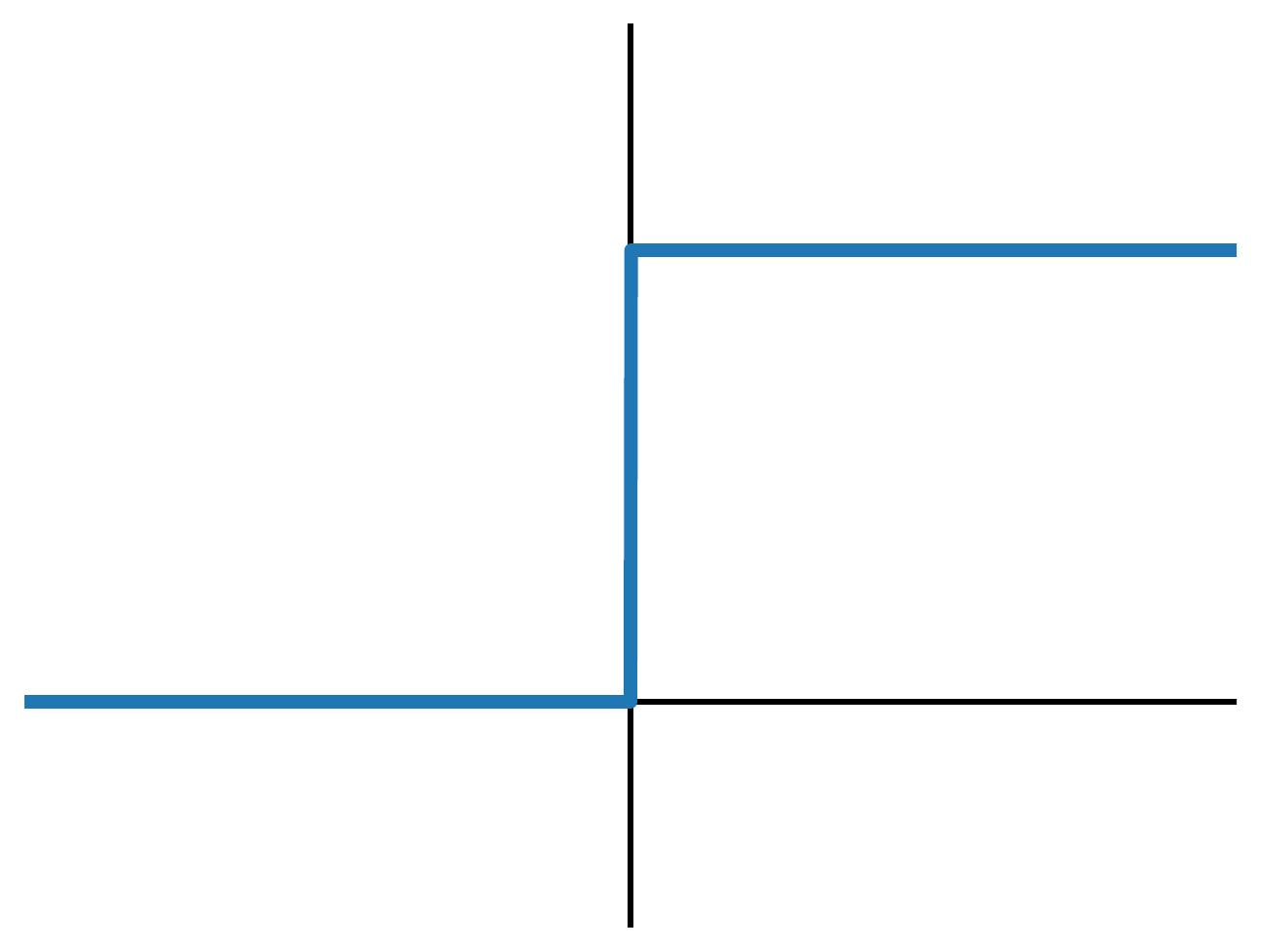}}
        (b) Unit Step
    \end{minipage}
    \begin{minipage}[b]{0.3\linewidth}
        \centering
        \centerline{\includegraphics[width=\textwidth]{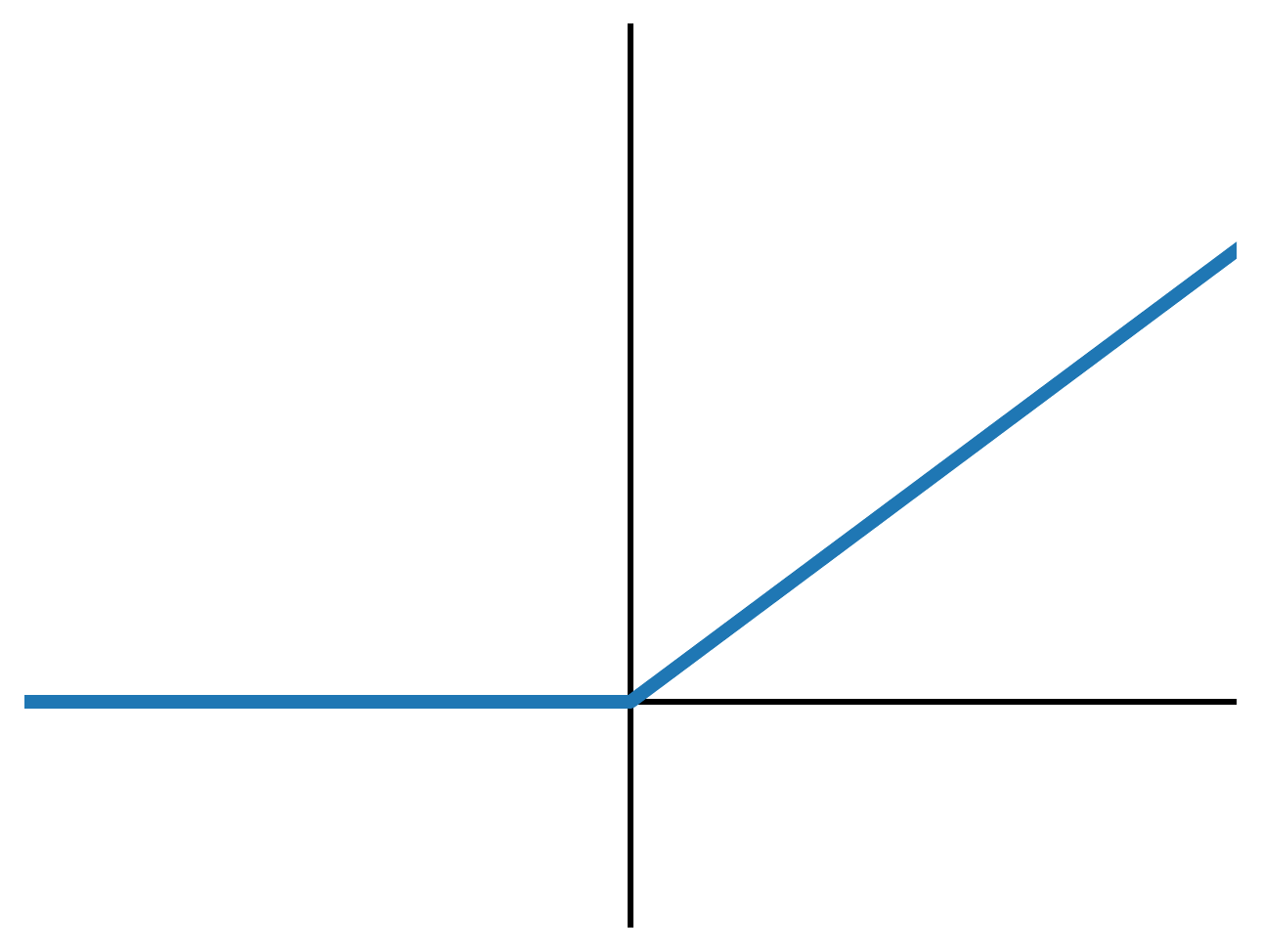}}
        (c) ReLU
    \end{minipage}
    \caption{Typical activation functions found in neural networks.}
    \label{fig:activations}
\end{figure}

The task of DL corresponds to learning the input-output mapping from a data set in a hierarchical or multi-layer manner. Deep neural networks (DNNs) are complicated function mappings built from many smaller, simpler building blocks. The simplest building block of a DNN is an (artificial) neuron, inspired by the biological neurons of the brain~\cite{McCullochAN}. A neuron is a function mapping $\R^d \to \R$ of the form $\vec{z} \mapsto \sigma(\vec{w}^\T\vec{z} - b)$, where $\vec{w} \in \R^d$ corresponds to the \emph{weights} of the neuron and $b \in \R$ corresponds to the bias of the neuron. The function $\sigma: \R \to \R$ is referred to as the \emph{activation function} of the neuron and controls nonlinear response of the neuron. A neuron ``activates'' when the weighted combination of its input $\vec{x}$ exceeds a certain threshold, i.e., $\vec{w}^\T\vec{x} > b$. Therefore, typical activation functions such as the sigmoid, unit step function, or rectified linear unit (ReLU) activate when their input exceeds $0$ as seen in \cref{fig:activations}.

\begin{figure}[t]
    \centering
    \begin{minipage}[b]{\linewidth}
        \centering
        \centerline{\includegraphics[width=0.8\textwidth]{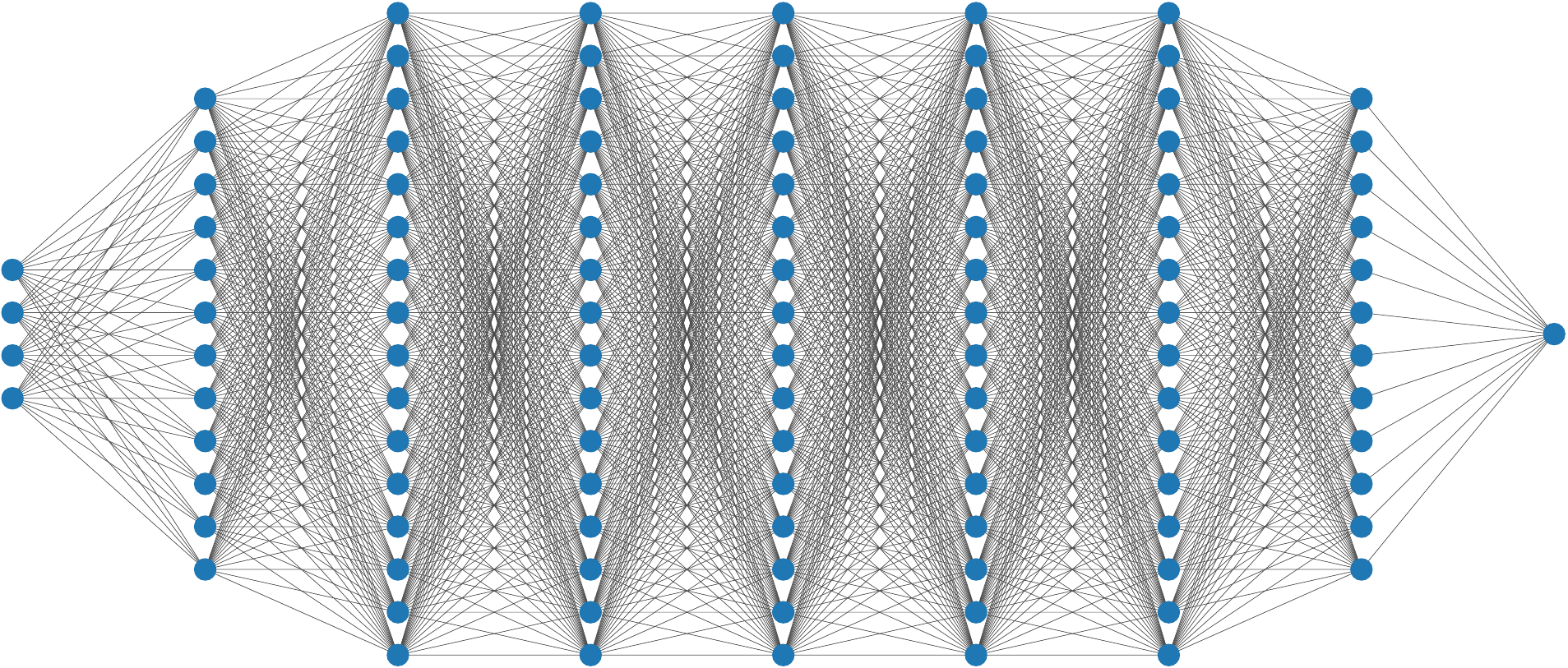}}
        (a) Feedforward DNN
    \end{minipage}
    
    \vspace{3em}
    
    \begin{minipage}[b]{\linewidth}
        \centering
        \centerline{\includegraphics[width=\textwidth]{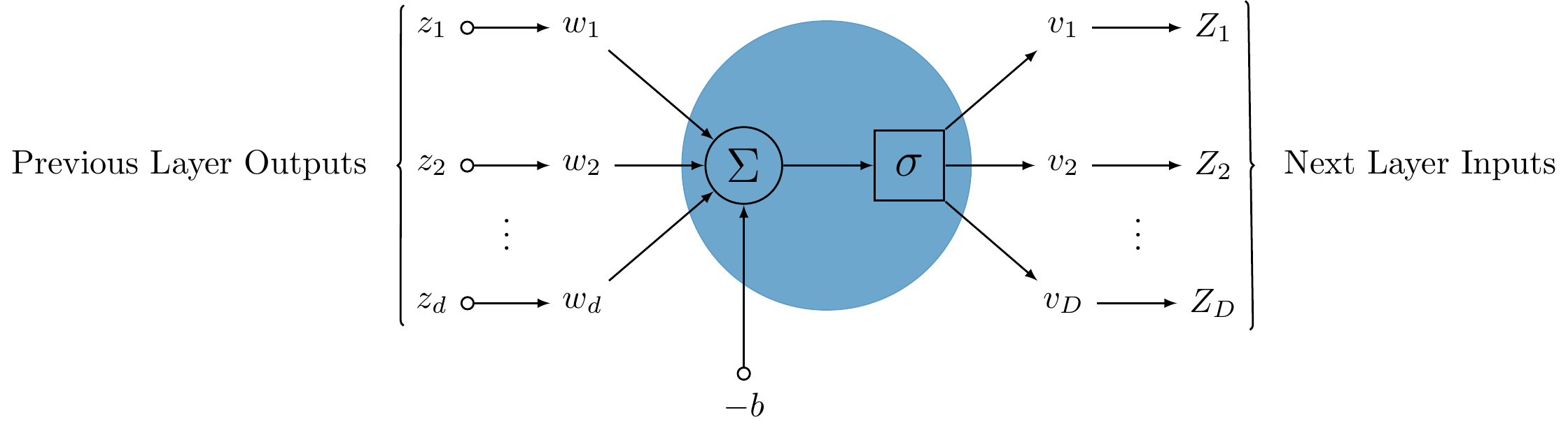}}
        (b) A Single Neuron
    \end{minipage}
    \caption{Example depiction of a deep neural network and its components: (a) a feedforward DNN architecture where the nodes represent the neurons and the edges represent the weights; (b) a single neuron from the DNN in (a) mapping an input $\vec{z} \in \R^d$ to an output $\vec{Z} \in \R^D$ via $\vec{Z} = \vec{v}\sigma(\vec{w}^\T\vec{z} - b).$}
    \label{fig:NN-diagrams}
\end{figure}

A neuron is composed of a linear mapping followed by a nonlinearity. A popular form (or ``architecture'') of a DNN is a fully-connected feedforward DNN which is a cascade of alternating linear mappings and component-wise nonlinearities. A feedforward DNN $f_\vec{\theta}$ (parameterized by $\vec{\theta}$) can be represented as the function composition
\[
    f_\vec{\theta}(\vec{x}) = \vec{A}^{(L)} \circ \vec{\sigma} \circ \vec{A}^{(L - 1)} \circ \cdots \vec{\sigma} \circ \vec{A}^{(1)}(\vec{x}),
    \label{eq:ff-DNN}
\]
where, for each $\ell = 1, \ldots, L$, the function $\vec{A}^{(\ell)}(\vec{z}) = \mat{W}^{(\ell)}\vec{z} - \vec{b}^{(\ell)}$ is an affine linear mapping with weight matrix $\mat{W}^{(\ell)}$ and bias vector $\vec{b}^{(\ell)}$. The functions $\vec{\sigma}$ that appear in the composition apply the activation function $\sigma: \R \to \R$ component-wise to the vector $\vec{A}^{(\ell)}(\vec{z})$.
While the activation function could change from neuron to neuron, we assume that the same activation function is used in the entire network in this article.
The parameters of this DNN are the weights and biases, i.e., $\vec{\theta} = \curly{(\mat{W}^{(\ell)}, \vec{b}^{(\ell)})}_{\ell=1}^L$. Each  mapping $\vec{A}^{(\ell)}$ corresponds to a \emph{layer} of the DNN and the number of  mappings $L$ is the \emph{depth} of the DNN. The dimensions of the weight matrices $\mat{W}^{(\ell)}$ correspond to the number of neurons in each layer (i.e., the \emph{width} of the layer). Alternative DNN architectures can be built with other simple building blocks, e.g., with convolutions and pooling/downsampling operations, which would correspond to deep convolutional neural networks (DCNNs). DNN architectures are often depicted with diagrams as in \cref{fig:NN-diagrams}.

Given a DNN $f_\vec{\theta}$ parameterized by $\vec{\theta} \in \Theta$ (of any architecture), the task of learning from the data $\curly{(\vec{x}_n, y_n)}_{n=1}^N$ is formulated as the optimization problem
\[
    \min_{\vec{\theta} \in \Theta} \sum_{n=1}^N \mathcal{L}(y_n, f_\vec{\theta}(\vec{x}_n)),
\]
where $\mathcal{L}(\dummy, \dummy)$ is a loss function (squared error, logistic, hinge loss, etc.). For example, the squared error loss is given by $\mathcal{L}(y, z)= (y-z)^2$. A DNN is \emph{trained} by solving this optimization problem, usually via some form of gradient descent.  In typical scenarios, this optimization problem is ill-posed and so the problem is regularized either explicitly through the addition of a regularization term and/or implicitly by constraints on the network architecture or the behavior of gradient descent procedures~\cite{PoggioDL}. A surprising phenomenon of gradient descent training algorithms for overparameterized NNs is that, among the many solutions which overfit the data, these algorithms select one which often generalizes well on new data, even without explicit regularization. This has led to researchers trying to understand the role of overparameterization and the effect of random initialization of NN parameters on the implicit bias of gradient-based training algorithms~\cite{ChizatImplicit}. 

On the other hand, explicit regularization corresponds to solving an optimization problem of the form
\[
    \min_{\vec{\theta} \in \Theta} \sum_{n=1}^N \mathcal{L}(y_n, f_\vec{\theta}(\vec{x}_n)) + \lambda C(\vec{\theta}),
\]
where $C(\vec{\theta})\geq 0$ for all $\vec{\theta}\in\Theta$. $C(\vec{\theta})$ is a \emph{regularizer} which measures the ``size'' (or ``capacity'') of the DNN parameterized by $\vec{\theta} \in \Theta$ and $\lambda > 0$ is an adjustable hyperparamter which controls a trade-off between the data-fitting term and the regularizer. DNNs are often trained using gradient descent algorithms with \emph{weight decay} which corresponds to solving the optimization problem
\[
    \min_{\vec{\theta} \in \Theta} \sum_{n=1}^N \mathcal{L}(y_n, f_\vec{\theta}(\vec{x}_n)) + \lambda C_{\mathrm{wd}}(\vec{\theta}),
    \label{eq:DNN-weight-decay}
\]
where the weight decay regularizer $C_\mathrm{wd}(\vec{\theta})$ is the squared Euclidean-norm of all the network weights. Sometimes the weight decay objective 
regularizes all parameters, including biases, while sometimes it only regularizes the weights (so that the biases are unregularized). This article focuses on the variant of weight decay with unregularized biases.
\section{What is the Effect of Regularization in Deep Learning?}
Weight decay is a common form of regularization for DNNs. On the surface, it appears to simply be the familiar Tikhonov or ``ridge'' regularization. In standard linear models, it is well-known that this sort of regularization tends to reduce the size of the weights, but does not produce sparse weights. However, when this regularization is used in conjunction with NNs, the results are strikingly different.  Regularizing the sum of squared weights turns out to be equivalent to regularization with a type of  $\ell^1$-norm  regularization on the network weights, leading to \emph{sparse} solutions in which the weights of many neurons are zero~\cite{YangWeightDecayProx}. This is due to the key property that the most commonly used activation functions in  DNNs are \emph{homogeneous}. A function $\sigma(t)$ is said to be homogeneous (of degree $1$) if $\sigma(\gamma t) = \gamma \sigma(t)$ for any $\gamma > 0$. The most common NN activation function, the ReLU, is homogeneous, as well as the leaky ReLU, the linear activation, and pooling/downsampling units.  This homogeneity leads to the following theorem, referred to as the \emph{neural balance theorem} (NBT). 

\begin{GrayBox}
\textbf{Neural Balance Theorem~(\cite[Theorem~1]{YangWeightDecayProx})}:
    Let $f_\vec{\theta}$ be a DNN of any architecture parameterized by $\vec{\theta} \in \Theta$ which solves the DNN training problem with weight decay in \cref{eq:DNN-weight-decay}. Then, the weights satisfy the following balance constraint: if $\vec{w}$ and $\vec{v}$ denote the input and output weights of any homogeneous unit in the DNN, then $\norm{\vec{w}}_2 = \norm{\vec{v}}_2$.
\end{GrayBox}

The proof of this theorem boils down to the simple observation that for any homogeneous unit with input weights $\vec{w}$ and output weights $\vec{v}$, we can scale the input weight by $\gamma > 0$ and the output weight by $1/\gamma$ without changing the function mapping. For example, consider the single neuron $\vec{z} \mapsto \vec{v}\sigma(\vec{w}^\T\vec{z} - b)$ with homogeneous activation function $\sigma$ as depicted in \cref{fig:NN-diagrams}(b). In the case of a DNN as in \cref{eq:ff-DNN}, $\vec{w}$ corresponds to a row of a weight matrix in the affine mapping of a layer, $\vec{v}$ corresponds to a column of the weight matrix in the subsequent layer, and $b$ corresponds to an entry in the bias vector. It is immediate that $(\vec{v} / \gamma) \sigma((\gamma\vec{w})^\T\vec{z} - \gamma b) = \vec{v}\sigma(\vec{w}^\T\vec{z} - b)$. By noting that the biases are unregularized, the theorem follows by noticing that $\min_{\gamma > 0} \norm{\gamma \vec{w}}_2^2 + \norm{\vec{v} / \gamma}_2^2$ occurs when $\gamma = \sqrt{\norm{\vec{v}}_2 / \norm{\vec{w}}_2}$ which implies that the minimum squared Euclidean-norm solution must satisfy the property that the input and output weights $\vec{w}$ and $\vec{v}$ are balanced.

\subsection{The Secret Sparsity of Weight Decay} \label{sec:secret-sparsity}
The balancing effect of the NBT has a striking effect on  solutions to the weight decay objective. In particular, a sparsity-promoting effect akin to least absolute shrinkage and selection operator (LASSO) regularization~\cite{TibshiraniLASSO}. As an illustrative example, consider a \emph{shallow} ($L = 2$), feedforward NN mapping $\R^d \to \R^D$ with a homogeneous activation function (e.g., the ReLU) and $K$ neurons. In this case, the NN is given by
\[
    f_\vec{\theta}(\vec{x}) = \sum_{k=1}^K \vec{v}_k \sigma(\vec{w}_k^\T\vec{x} - b_k).
    \label{eq:shallow-multivariate-vv}
\]
Here, the weight decay regularizer is of the form $\frac{1}{2}\sum_{k=1}^{K} \norm{\vec{v}_k}_2^2 + \norm{\vec{w}_k}_2^2$, where $\vec{w}_k$ and $\vec{v}_k$ are the input and output weights of the $k$th neuron, respectively. By the NBT, this is equivalent to using the regularizer $\sum_{k=1}^{K} \norm{\vec{v}_k}_2 \norm{\vec{w}_k}_2$. Due to the homogeneity of the activation function, we can assume, without loss of generality, that $\norm{\vec{w}_k}_2=1$ by ``absorbing'' the magnitude of the input weight $\vec{w}_k$ into the output weight $\vec{v}_k$. Therefore, by constraining the input weights to be unit-norm, the training problem can then be reformulated with the regularizer $\sum_{k=1}^{K} \norm{\vec{v}_k}_2$~\cite{YangWeightDecayProx}. Remarkably, this is exactly the well-known group LASSO regularizer~\cite{YuanGLASSO}, which favors solutions with few active neuron connections (i.e., solutions typically have many $\vec{v}_k$ exactly equal to $\vec{0}$), although the overall training objective remains nonconvex. We also note that there is a line of work that has reformulated the nonconvex training problem as a convex group LASSO problem~\cite{PilanciConvex}.

More generally, consider the feedforward \emph{deep} NN architecture in \cref{eq:ff-DNN} with a homogeneous activation function and consider training the DNN with weight decay only on the network weights. An application of the NBT shows that the weight decay problem is equivalent to the regularized DNN training problem with the regularizer
\[
    C(\vec{\theta}) = \frac{1}{2} \sum_{k=1}^{K^{(1)}} \norm{\vec{w}^{(1)}_k}_2^2 + \frac{1}{2} \sum_{k=1}^{K^{(L)}} \norm{\vec{v}^{(L)}_k}_2^2 + \sum_{\ell=1}^L \sum_{k=1}^{K^{(\ell)}} \norm{\vec{w}^{(\ell)}_k}_2 \norm{\vec{v}^{(\ell)}_k}_2,
    \label{eq:ff-DNN-sum-of-paths}
\]
where $K^{(\ell)}$ denotes the number of neurons in layer $\ell$, $\vec{w}_k^{(\ell)}$ denotes the input weights to the $k$th neuron in layer $\ell$, and $\vec{v}_k^{(\ell)}$ denotes the output weights to the $k$th neuron in layer $\ell$ (see~\cite[Equation~(2)]{YangWeightDecayProx}). Solutions based on this regularizer will also be sparse due to the $2$-norms that appear in the last term in \cref{eq:ff-DNN-sum-of-paths} being \emph{not squared}, akin to the group LASSO regularizer. In particular, this regularizer can be viewed as a mixed $\ell^{2,1}$-norm on the weight matrices. Moreover, increasing the regularization parameter $\lambda$, will increase the number of weights that are zero in the solution.  Therefore, training the DNN with weight decay favors sparse solutions, where sparsity is quantified via the number of active neuron connections. An early version of this result appeared in 1998~\cite{Grandvalet}, although it did not become well-known until it was rediscovered in 2015~\cite{NeyshaburInductiveBias}.

\section{What Kinds of Functions Do Neural Networks Learn?} \label{sec:kinds}
The sparsity-promoting effect of weight decay arising from the NBT in network architectures with homogeneous activation functions has several consequences on the properties of trained NNs. In this section, we will focus on the popular ReLU activation function, $\rho(t) = \max\curly{0, t}$. The imposed sparsity not only promotes sparsity in the sense of the number of active neuron connections, but also promotes a kind of \emph{transform-domain} sparsity. This transform-domain sparsity suggests the inclusion of skip connections and low-rank weight matrices in network architectures.

\subsection{Shallow Neural Networks}
In the univariate case, a shallow, feedforward ReLU NN with $K$ neurons is realized by the mapping
\[
    f_\vec{\theta}(x) = \sum_{k=1}^K v_k \rho(w_k x - b_k).
    \label{eq:shallow-univariate}
\]
Training this NN with weight decay corresponds to the solving the optimization problem
\[
    \min_{\vec{\theta} \in \Theta} \sum_{n=1}^N \mathcal{L}(y_n, f_\vec{\theta}(x_n)) + \frac{\lambda}{2} \sum_{k=1}^K \abs{v_k}^2 + \abs{w_k}^2,
    \label{eq:univariate-weight-decay}
\]
From \cref{sec:secret-sparsity}, we saw that the NBT implies that this problem is equivalent to
\[
    \min_{\vec{\theta} \in \Theta} \sum_{n=1}^N \mathcal{L}(y_n, f_\vec{\theta}(x_n)) + \lambda \sum_{k=1}^K \abs{v_k}\abs{w_k}.
    \label{eq:univariate-path-norm}
\]
\begin{tcolorbox}[float*=t,
    width=\linewidth,
    toprule = 1mm,
    bottomrule = 0.5mm,
    leftrule = 0.5mm,
    rightrule = 0.5mm,
    arc = 0mm,
    fonttitle = \sffamily\bfseries\large,
    title = {[IN1] ReLU Sparsity in the Second Derivative Domain}]
    \begin{multicols}{2}
    Given a ReLU neuron $r(x) = \rho(wx - b)$, its first derivative, $\D r(x)$, is
    \begin{align}
        \D r(x) &= \D \rho(wx - b) \nonumber \\
        &= w \, u(wx - b),
    \end{align}
    where $u$ is the unit step function (\cref{fig:activations}(b)). Therefore, its second derivative, $\D^2 r(x)$, is
    \begin{align}
        \D^2 r(x) &= \D w \, u(wx - b) \nonumber \\
        &= w^2 \, \delta(wx - b).
    \end{align}
    By the scaling property of the Dirac impulse~\cite[Problem~1.38(a)]{SS}
    \[
        \delta(\gamma x) = \frac{1}{\abs{\gamma}} \delta(x)
        \label{eq:Dirac-scaling}
    \]
    we have
    \begin{align}
        \D^2 r(x) &= \frac{w^2}{\abs{w}} \delta\paren*{x - \frac{b}{w}} \nonumber \\
        &= \abs{w} \delta\paren*{x - \frac{b}{w}}.
        \label{eq:univariate-sparsified}
    \end{align}
    The second derivative of the NN \cref{eq:shallow-univariate} is then
    \[
        \D^2 f_\vec{\theta}(x) = \sum_{k=1}^K v_k \abs{w_k} \delta\paren*{x - \frac{b_k}{w_k}}.
    \]
    Therefore,
    \[
        \int_{-\infty}^\infty \abs{\D^2 f_\vec{\theta}(x)} \dd x = \sum_{k=1}^K \abs{v_k}\abs{w_k}.
        \label{eq:integral-path-norm}
    \]
    \end{multicols}

    \vspace{2em}
    \hrule
    \vspace{2em}

    \begin{minipage}[t]{\textwidth}
    \centering
    \includegraphics[width=0.3\textwidth]{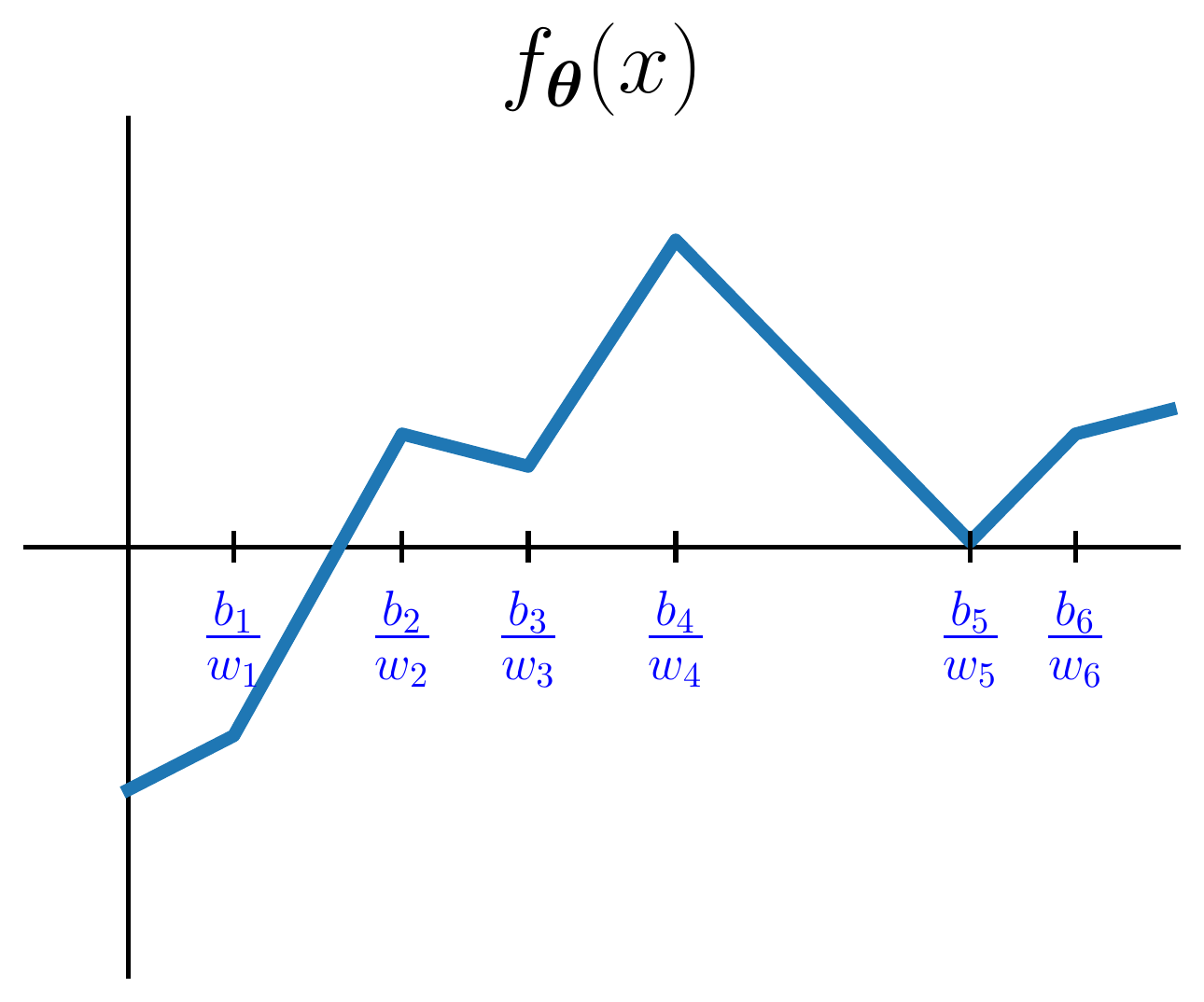}
    \includegraphics[width=0.3\textwidth]{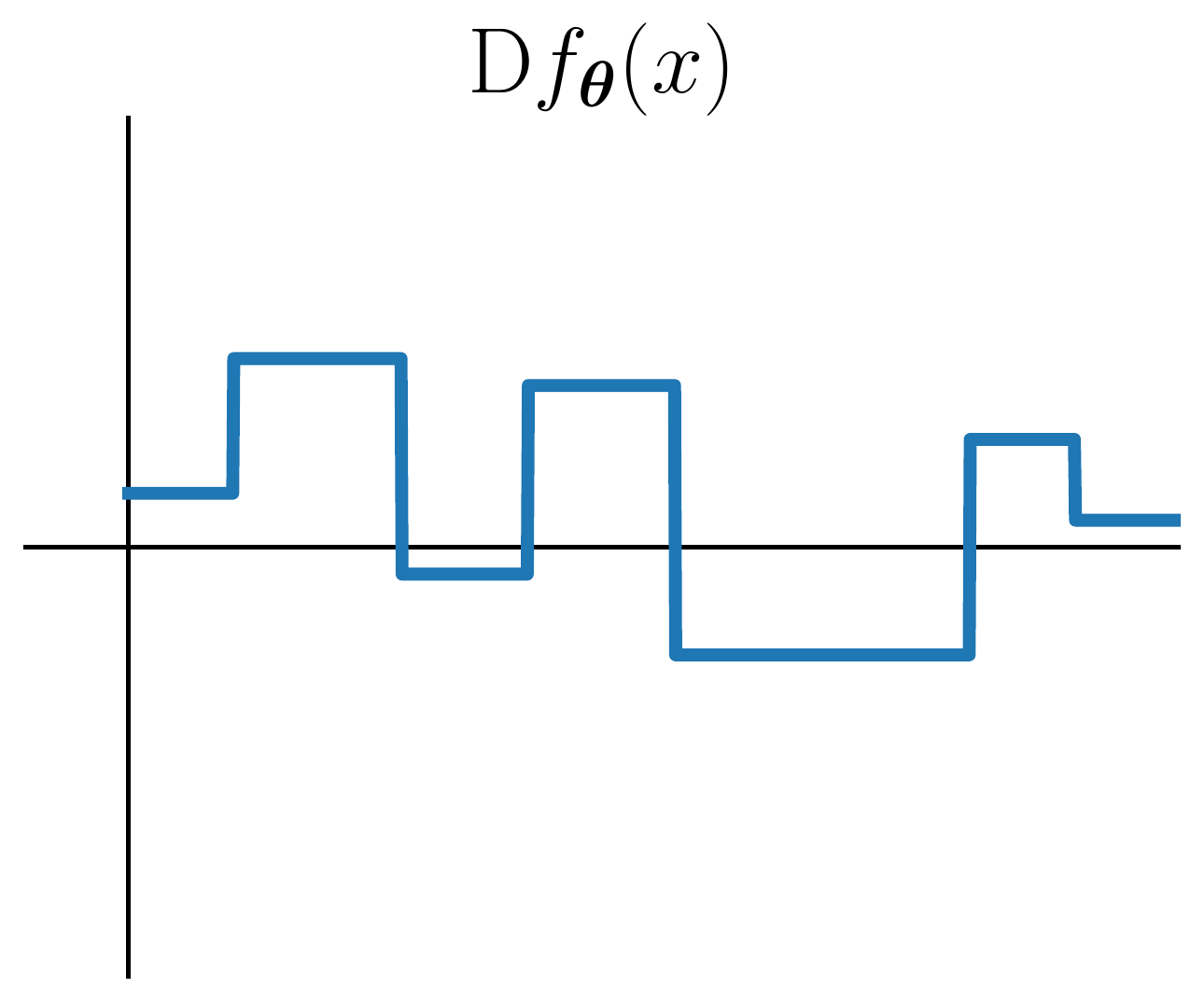}
    \includegraphics[width=0.3\textwidth]{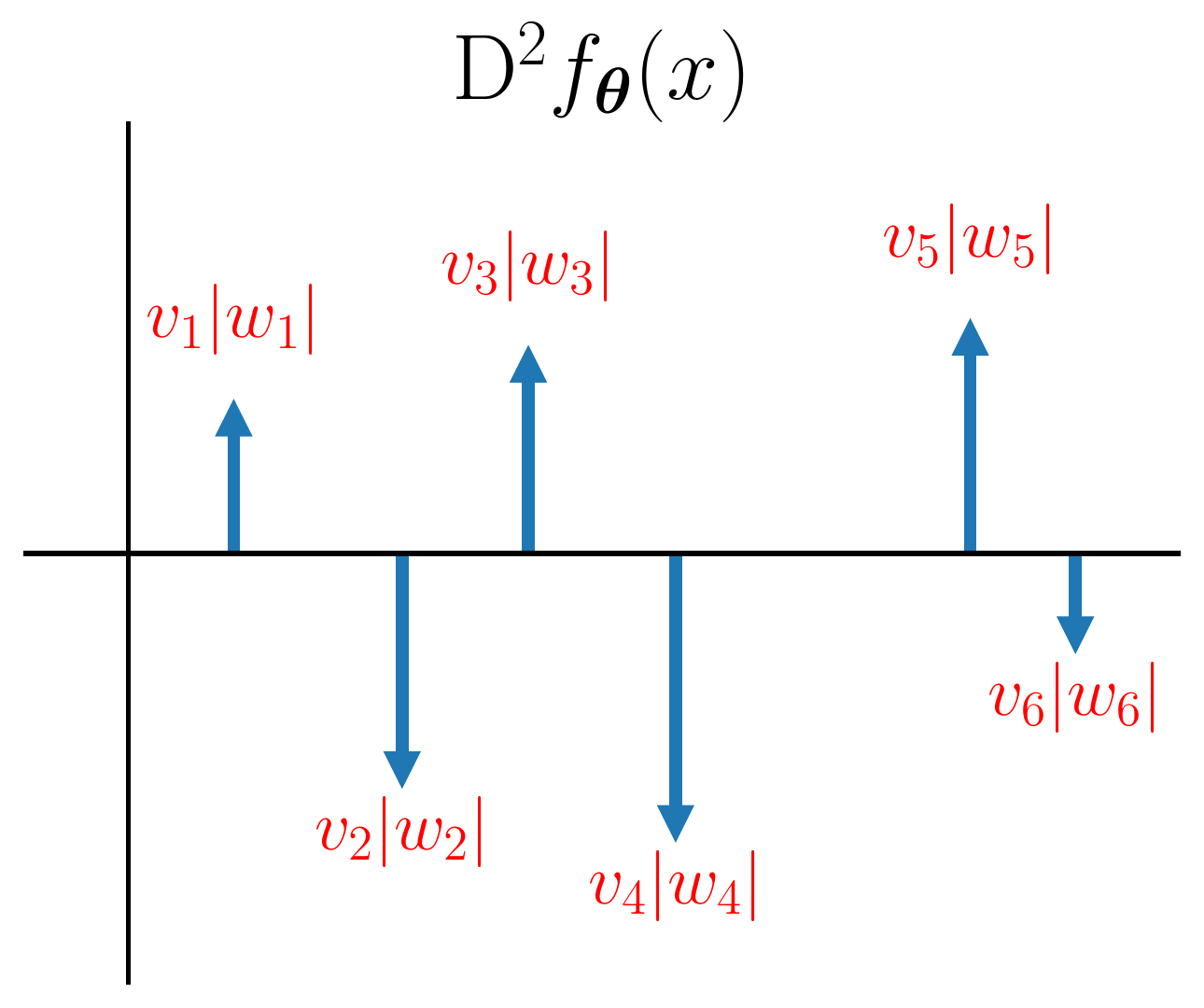}
    \captionof{figure}{Illustration of the sparsity in the second derivative domain of a univariate, shallow feedforward NN with $6$ neurons.}
    \label{fig:D2-sparsity}
    \end{minipage}
\end{tcolorbox}%
As illustrated in Insert IN1, we see that \cref{eq:univariate-path-norm} is actually regularizing the integral of second derivative of the NN, which can be viewed as a measure of sparsity in the second derivative domain. The integral in \cref{eq:integral-path-norm} must be understood in the \emph{distributional sense} since the Dirac impulse is not a function, but a generalized function or \emph{distribution}. To make this precise, let $g_\varepsilon(x) = e^{-x^2/2\varepsilon} / \sqrt{2\pi\varepsilon}$ denote the Gaussian density with variance $\varepsilon > 0$. As is well-known in signal processing, $g_\varepsilon$ converges to the Dirac impulse as $\varepsilon \to 0$. Using this idea, given a distribution $f$, define the norm
\[
    \norm{f}_\M \coloneqq \sup_{\varepsilon > 0} \: \norm{f * g_\varepsilon}_{L^1} = \sup_{\varepsilon > 0} \int_{-\infty}^\infty \abs*{\int_{-\infty}^\infty f(x) g_\varepsilon(y - x) \dd x} \dd y.
\]
This definition provides an explicit construction, via the convolution with a Gaussian, of a sequence of smooth functions that converge to $f$, where the supremum acts as the limit. For example, if $f(x) = g(x) + \sum_{k=1}^K v_k \delta(x - t_k)$, where $g$ is an absolutely integrable function, then $\norm{f}_\M = \norm{g}_{L^1} + \sum_{k=1}^K \abs{v_k} = \norm{g}_{L^1} + \norm{\vec{v}}_1$, with $\norm{\vec{v}}_1= \sum_{k=1}^K |v_k|$. It is in this sense that \cref{eq:integral-path-norm} holds, i.e., $\norm{\D^2 f_\vec{\theta}}_\M = \sum_{k=1}^K \abs{v_k} \abs{w_k}$. In particular, the $\M$-norm is precisely the continuous-domain analogue of the sparsity-promoting discrete $\ell^1$-norm. Therefore, we see that training a NN with weight decay as in \cref{eq:univariate-weight-decay} prefers solutions with sparse second derivatives.

It turns out that the connection between between sparsity in the second derivative domain and NNs is even tighter. Let $\BV^2(\R)$ denote the space of functions mapping $\R \to \R$ such that $\norm{\D^2 f}_\M$ is finite. This is the space of functions of second-order \emph{bounded variation} and the quantity $\norm{\D^2 f}_\M$ is the second-order \emph{total variation} (TV)\footnote{The classical notion of TV, often used in signal denoising problems~\cite{RudinTV}, is $\TV(f) \coloneqq \norm{\D f}_\M$ and so the second-order TV of $f$ can be viewed as the TV of the derivative of $f$: $\norm{\D^2 f}_\M = \TV(\D f)$.} of $f$. 
It is well-known from spline theory~\cite{FisherJerome,MammenLAS,UnserSplinesgTV} that functions that fit data and have minimimal second-order TV are continuous piecewise linear (CPwL) functions. Since the ReLU is a CPwL function,  ReLU NNs are CPwL functions~\cite{BalestrieroMadMax}. In fact, under mild assumptions on the loss function, the solution set to the optimization problem
\[
    \min_{f \in \BV^2(\R)} \sum_{n=1}^N \mathcal{L}(y_n, f(x_n)) + \lambda \norm{\D^2 f}_\M
    \label{eq:BV2-problem}
\]
is completely characterized by NNs of the form
\[
    f_\vec{\theta}(x) = \sum_{k=1}^K v_k \rho(w_k x - b_k) + c_1 x + c_0,
    \label{eq:shallow-univariate-skip}
\]
where the number of neurons is strictly less than the number of data ($K < N$)
in the sense that the solution set to \cref{eq:BV2-problem} is a closed convex set whose extreme points take the form of \cref{eq:shallow-univariate-skip} with $K < N$~\cite{DebarreSparsest,ParhiSPL2020,SavareseInfiniteWidth}.
In neural network parlance, the $c_1x + c_0$ term is a \emph{skip connection}~\cite{HeResidual}. This term is an affine function that naturally arises since the second-order TV of an affine function is zero and so the regularizer places no penalty for the inclusion of this term.

The intuition behind this result is due the fact that the second derivative of a CPwL function is an impulse train and therefore exhibits extreme sparsity in the second derivative domain. This is illustrated in \cref{fig:D2-sparsity}. Therefore, the optimization problem \cref{eq:BV2-problem} will favor sparse CPwL functions which always admit a representation as in \cref{eq:shallow-univariate-skip}. In signal processing parlance, ``signals'' that are sparse in some transform domain are said to have a \emph{finite rate of innovation}~\cite{VetterliFRI}. Here, the transform involved is the second derivative operator and the innovation is the impulse train that arises after applying the second derivative operator to a CPwL function.

Consider the optimization over the NN parameter space $\Theta_K$ of networks as in \cref{eq:shallow-univariate-skip} with fixed width $K \geq N$. From the derivation in Insert IN1, we have $\curly{f_\vec{\theta} \st \vec{\theta} \in \Theta_K} \subset \BV^2(\R)$. Furthermore, from the above discussion, we know there always exists an optimal solution to \cref{eq:BV2-problem} that takes the form of \cref{eq:shallow-univariate-skip} with $K < N$, i.e., there always exists a solution to \cref{eq:BV2-problem} in $\curly{f_\vec{\theta} \st \vec{\theta} \in \Theta_K}$. Therefore, from the equivalence of \cref{eq:univariate-weight-decay,eq:univariate-path-norm}, we see that training a sufficiently wide ($K \geq N$) NN with a skip connection \cref{eq:shallow-univariate-skip} and weight decay \cref{eq:univariate-weight-decay} results in a solution to the optimization problem \cref{eq:BV2-problem} over the function space $\BV^2(\R)$. While this result may seem obvious in hindsight, it is remarkable since it says that the kinds of functions that NNs trained with weight decay (to a global minimizer) are \emph{exactly} optimal functions in $\BV^2(\R)$. Moreover, this result sheds light on the role of overparameterization since this correspondence hinges on the network being critically parameterized or overparameterized (because $K \geq N$).

\begin{tcolorbox}[float*=htb,
    width=\linewidth,
    toprule = 1mm,
    bottomrule = 0.5mm,
    leftrule = 0.5mm,
    rightrule = 0.5mm,
    arc = 0mm,
    fonttitle = \sffamily\bfseries\large,
    title = {[IN2] The Radon Transform and Fourier Slice Theorem}]
    \small
    \begin{multicols}{2}
        The Radon transform, first studied by Radon in 1917~\cite{Radon}, of a function  mapping $\R^d \to \R$ is specified by the integral transform
        \[
            \RadonOp\curly{f}(\vec{\alpha}, t) = \int_{\R^d} f(\vec{x}) \delta(\vec{\alpha}^\T\vec{x} - t) \dd\vec{x},
        \]
        where $\delta$ is the univariate Dirac impulse, $\vec{\alpha} \in \Sph^{d-1} = \curly{\vec{u} \in \R^d \st \norm{\vec{u}}_2 = 1}$ is a unit vector, and $t \in \R$ is a scalar. The Radon transform of $f$ at $(\vec{\alpha}, t)$ is the integral of $f$ along the hyperplane $\curly{\vec{x} \in \R^d \st \vec{\alpha}^\T\vec{x} = t}$.
        
        The Radon transform is tightly linked with the Fourier transform, specified by
        \[
        \hat{f}(\vec{\omega}) = \int_{\R^d} f(\vec{x}) e^{-\imag \vec{\omega}^\T\vec{x}} \dd\vec{x},
        \]
        where $\imag^2 = -1$. Indeed,
        \begin{align*}
            &\phantom{{}={}} \hat{\RadonOp\curly{f}}(\vec{\alpha}, \omega) \\
            &= \int_{\R} \paren*{\int_{\R^d} f(\vec{x}) \delta(\vec{\alpha}^\T\vec{x} - t) \dd\vec{x}} e^{-\imag \omega t} \dd t \\
            &= \int_{\R^d} f(\vec{x}) \paren*{\int_{\R} \delta(\vec{\alpha}^\T\vec{x} - t) e^{-\imag \omega t} \dd t} \dd\vec{x} \\
            &= \int_{\R^d} f(\vec{x}) e^{-\imag (\omega \vec{\alpha})^\T\vec{x}} \dd\vec{x} = \hat{f}(\omega\vec{\alpha}). \numberthis
        \end{align*}
        This result is known as the \emph{Fourier slice theorem}. It states that the univariate Fourier transform in the Radon domain corresponds to a slice of the Fourier transform in the spatial domain.
    \end{multicols}
\end{tcolorbox}

In the multivariate case, a shallow feedforward NN has the form
\[
    f_\vec{\theta}(\vec{x}) = \sum_{k=1}^K v_k \rho(\vec{w}_k^\T\vec{x} - b_k).
    \label{eq:shallow-multivariate}
\]
The key property connects the univariate case and $\BV^2(\R)$ was that ReLU neurons are \emph{sparsified} by the second derivative operator as in \cref{eq:univariate-sparsified}. A similar analysis can be carried out in the multivariate case by finding an operator that is the sparsifying transform of the multivariate ReLU neuron $r(\vec{x}) = \rho(\vec{w}^\T\vec{x} - b)$. The sparsifying transform was proposed in 2020 in the seminal work of Ongie et al.~\cite{OngieRadon}, and hinges on the Radon transform that arises in tomographic imaging. Connections between the Radon transform and neurons have been known since at least the 1990s, gaining popularity due to the proposal of ridgelets~\cite{CandesPhD} and early versions of the sparsifying transform for neurons were studied as early as 1997~\cite{KurkovaEstimates}. A summary of the Radon transform appears in IN2. The sparsifying transform for multivariate ReLU neurons is based on a result regarding the (filtered) Radon transform, which appears in Insert IN3.

\begin{tcolorbox}[float*=t,
    width=\linewidth,
    toprule = 1mm,
    bottomrule = 0.5mm,
    leftrule = 0.5mm,
    rightrule = 0.5mm,
    arc = 0mm,
    fonttitle = \sffamily\bfseries\large,
    title = {[IN3] Filtered Radon Transform of a Neuron with Unit-Norm Input Weights}]
    \small
    \begin{multicols}{2}
        First consider the neuron $r(\vec{x}) = \sigma(\vec{w}^\T\vec{x})$ with $\vec{w} = \vec{e}_1 = (1, 0, \ldots, 0)$ (the first canonical unit vector). In this case, $r(\vec{x}) = \sigma(x_1)$. By noticing that this function can be written as a tensor product, the Fourier transform is given by the following product 
        \[
            \hat{r}(\vec{\omega}) = \hat{\sigma}(\omega_1) \prod_{k=2}^d 2 \pi \delta(\omega_k).
        \]
        By the Fourier slice theorem,
        \[
            \hat{\RadonOp\curly{r}}(\vec{\alpha}, \omega) = \hat{\sigma}(\omega \alpha_1) \prod_{k=2}^d 2\pi \delta(\omega \alpha_k).
        \]
        By the scaling property of the Dirac impulse \cref{eq:Dirac-scaling}, the above quantity equals
        \[
           = \hat{\sigma}(\omega \alpha_1)\frac{(2\pi)^{d-1}}{\abs{\omega}^{d-1}} \delta(\alpha_2, \ldots, \alpha_d).
           \label{eq:why-K}
        \]
        If we define the filter via the frequency response
        \[
            \hat{\KOp f}(\omega) = \frac{\abs{\omega}^{d-1}}{2(2\pi)^{d-1}} \hat{f}(\omega),
            \label{eq:K}
        \]
        we find
        \[
            \reallywidehat{\KOp \RadonOp\curly{r}}(\vec{\alpha}, \omega) = \frac{\hat{\sigma}(\omega \alpha_1)}{2} \delta(\alpha_2, \ldots, \alpha_d).
        \]
        Taking the inverse Fourier transform,
        \begin{align}
            &\phantom{{}={}} \KOp \RadonOp\curly{r}(\vec{\alpha}, t) \nonumber \\
            &= \frac{1}{2\abs{\alpha_1}} \sigma\paren*{\frac{t}{\alpha_1}} \delta(\alpha_2, \ldots, \alpha_d) \nonumber \\
            &= \frac{\sigma(t) \delta(\alpha_1 - 1) + \sigma(-t)\delta(\alpha_1 + 1)}{2} \delta(\alpha_2, \ldots, \alpha_d) \nonumber \\
            &= \frac{\sigma(t) \delta(\vec{\alpha} - \vec{e}_1) + \sigma(-t)\delta(\vec{\alpha} + \vec{e}_1)}{2} \nonumber \\
            &\eqqcolon \P_\even\curly{\sigma(t) \delta(\vec{\alpha} - \vec{e}_1)},
        \end{align}
        where $\P_\even$ is the projector which extracts the even part of its input (in terms of the variables $(\vec{\alpha}, t)$). The second line holds by the dilation property of the Fourier transform~\cite[Equation (4.34)]{SS}
        \[
            \frac{1}{\abs{\gamma}} f\paren*{\frac{t}{\gamma}} \xleftrightarrow{\:\FourierOp\:} \hat{f}(\gamma \omega).
        \]
        Since $\vec{\alpha} \in \Sph^{d-1}$, the third line holds by observing that when $\alpha_1 = \pm 1$, $\alpha_2, \ldots, \alpha_d = 0$ and so the second line is $\sigma(\pm t) / 2$ multiplied by an impulse and when $\alpha_1 \neq \pm 1$, the second line is $0$, which is exactly the third line. By the rotation properties of the Fourier transform, we have the following result for the neuron $r(\vec{x}) = \sigma(\vec{w}^\T\vec{x})$
        \[
            \KOp \RadonOp\curly{r}(\vec{\alpha}, t) = \P_\even\curly{\sigma(t)\delta(\vec{\alpha} - \vec{w})},
            \label{eq:filtered-Radon-neuron}
        \]
        where $\vec{w} \in \Sph^{d-1}$.
    \end{multicols}
\end{tcolorbox}

The filter $\KOp$ in \cref{eq:K} is exactly the \emph{backprojection filter} that arises in the filtered backprojection algorithm in tomographic image reconstruction and acts as a high-pass filter (or ramp filter) to correct the attenuation of high frequencies from the Radon transform. The intuition behind this is that the Radon transform integrates a function along hyperplanes. In the univariate case, the magnitude of the frequency response of an integrator behaves as $1 / \abs{\omega}$ and therefore attenuates high frequencies. The magnitude of the frequency response of integration along a hyperplane, therefore, behaves as $1 / \abs{\omega}^{d-1}$, since hyperplanes are of dimension $(d-1)$. Note that the even projector that appears in \cref{eq:filtered-Radon-neuron} is due to the fact that the Radon-domain variables $(\vec{\alpha}, t)$ and $(-\vec{\alpha}, -t)$ parameterize the same hyperplane.

\begin{figure}[t]
    \centering
    \begin{minipage}[b]{0.45\linewidth}
        \centering
        \centerline{\includegraphics[width=0.8\textwidth]{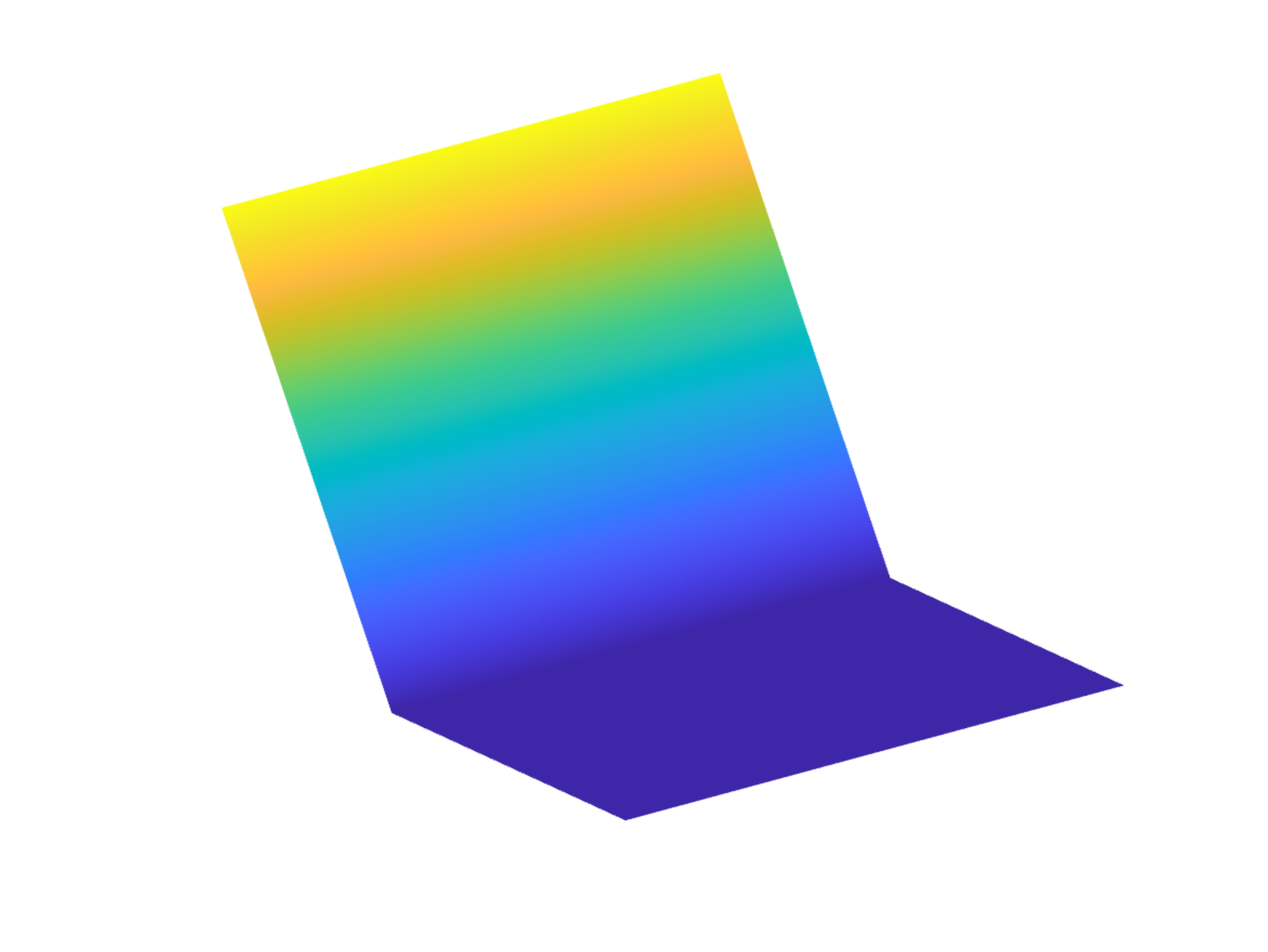}}
        \vspace{2em}
        (a) Surface plot of $r(\vec{x})$.
    \end{minipage}
    \begin{minipage}[b]{0.45\linewidth}
        \centering
        \centerline{\includegraphics[width=\textwidth]{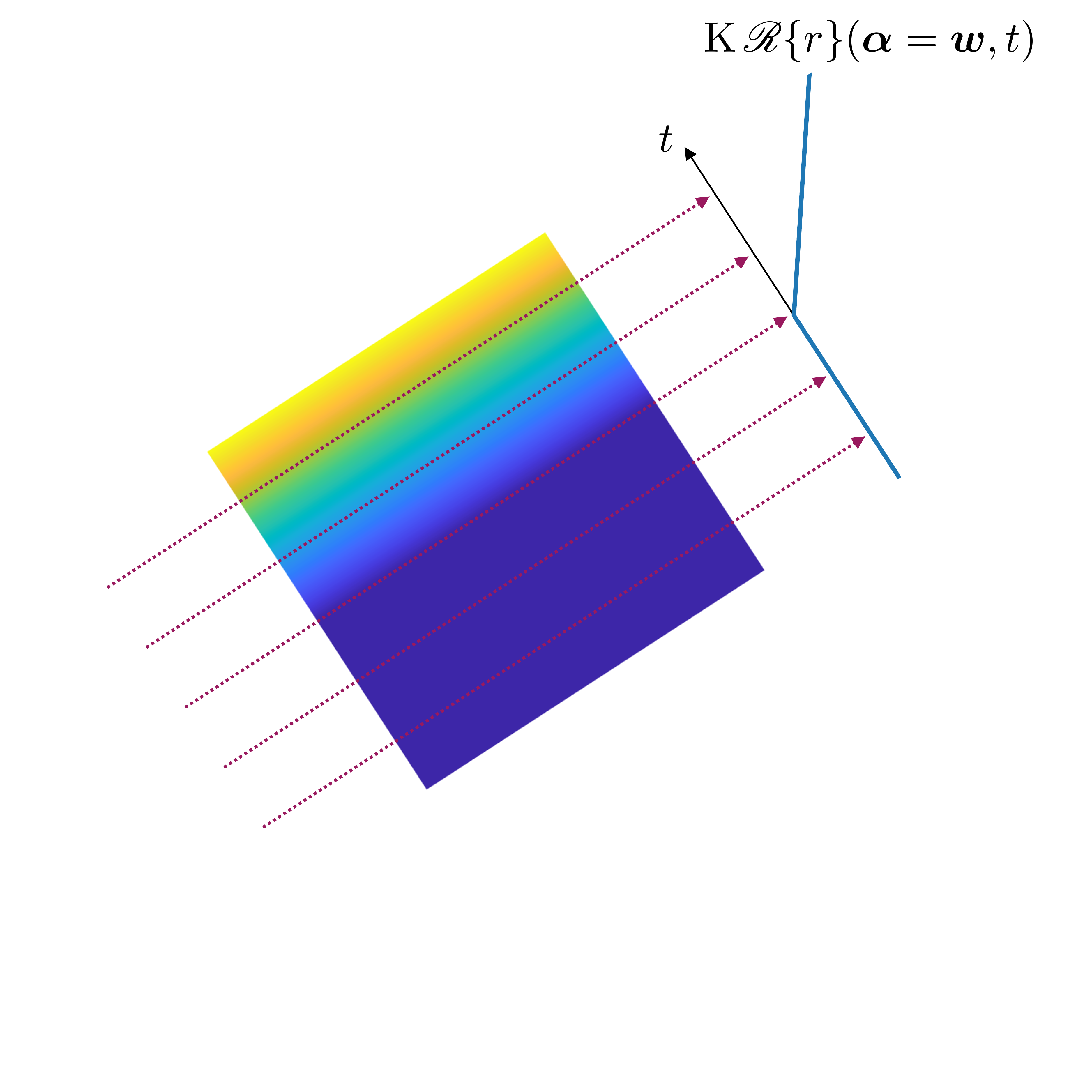}}
        (b) Filtered Radon transform when $\vec{\alpha} = \vec{w}$.
    \end{minipage}
    \begin{minipage}[b]{0.45\linewidth}
        \centering
        \centerline{\includegraphics[width=\textwidth]{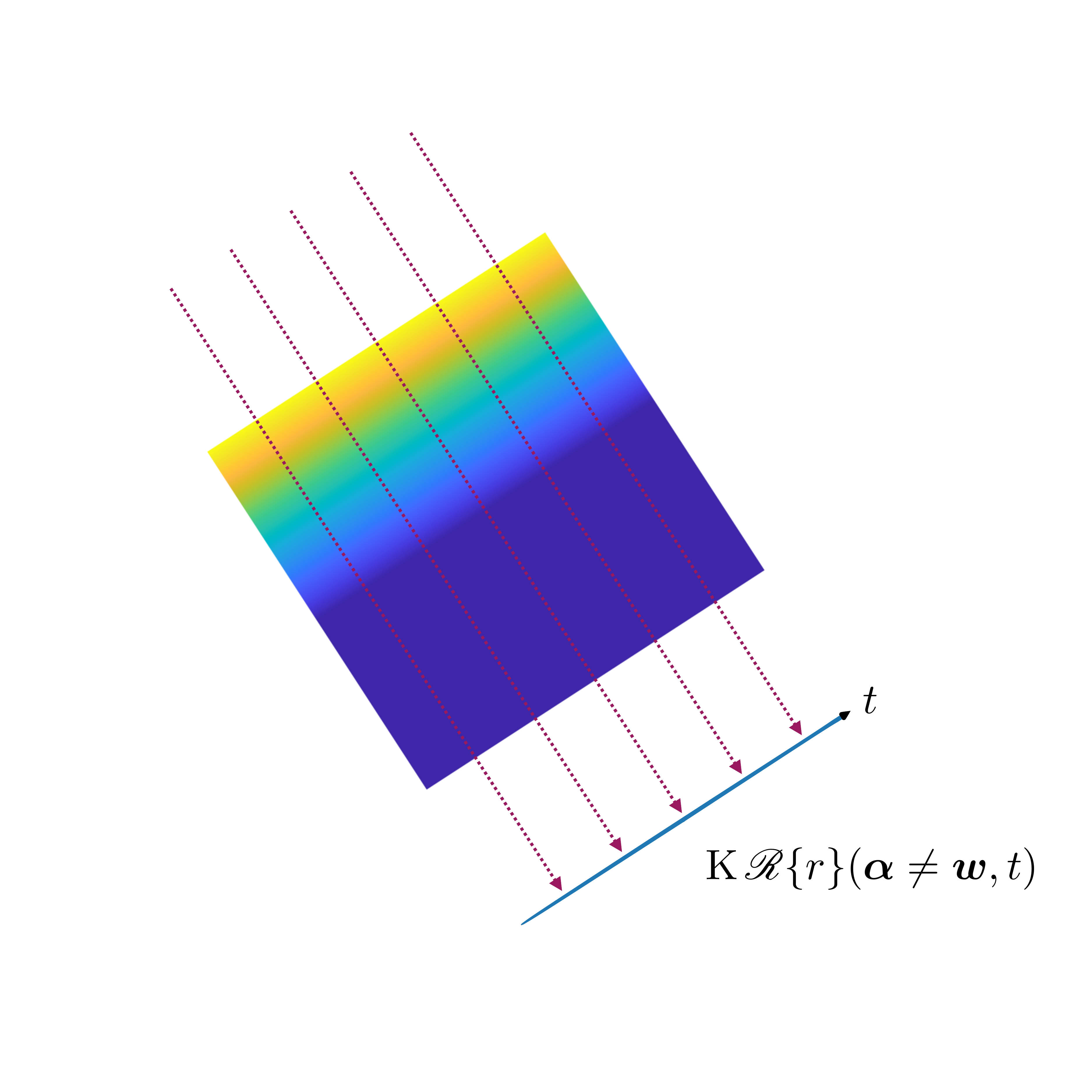}}
        (c) Filtered Radon transform when $\vec{\alpha} \neq \vec{w}$.
    \end{minipage}
    \begin{minipage}[b]{0.45\linewidth}
        \centering
        \centerline{\includegraphics[width=\textwidth]{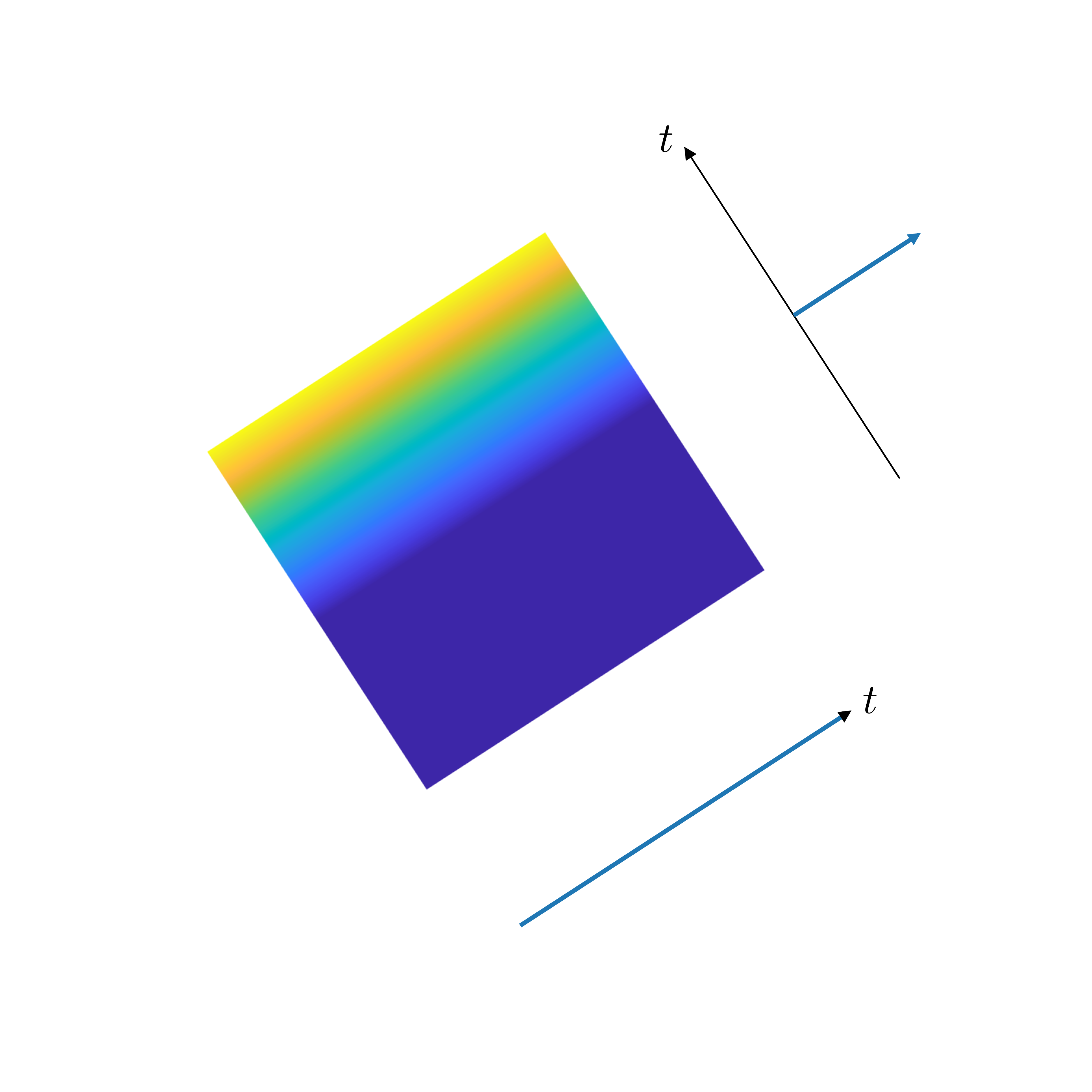}}
        (d) Sparsifying transform $\D_t^2\KOp \RadonOp \curly{r}$.
    \end{minipage}
    \caption{Cartoon diagram illustrating the illustrating the sparsifying transform of the ReLU neuron $r(\vec{x}) = \rho(\vec{w}^\T\vec{x} - b)$ with $(\vec{w}, b) \in \cyl$. The heatmap is a top-down view of a ReLU neuron depicted in (a).}
    \label{fig:Radon-ReLU}
\end{figure}

From the derivation in Insert IN3, we immediately see that the sparsifying transform of the multivariate ReLU neuron $r(\vec{x}) = \rho(\vec{w}^\T\vec{x} - b)$ with $(\vec{w}, b) \in \cyl$ is the operator $\D_t^2 \KOp \RadonOp$, where $\D_t^2 = \partial^2/\partial t^2$ denotes the second-order partial derivative with respect to $t$. We have
\[
    \D_t^2 \KOp \RadonOp \curly{r}(\vec{\alpha}, t) = \delta_{\RadonOp}((\vec{\alpha}, t) - (\vec{w}, b)),
    \label{eq:sparsifying-transform}
\]
where $\delta_{\RadonOp}(\vec{z} - \vec{z}_0) \coloneqq \P_\even\curly{\delta(\vec{z} - \vec{z}_0)} = (\delta(\vec{z} - \vec{z}_0) + \delta(\vec{z} + \vec{z}_0)) / 2$ is the even symmetrization of the Dirac impulse that arises due to the even symmetry of the Radon domain. From the homogeneity of the ReLU activation, applying this sparsifying transform to the (unconstrained) neuron $r(\vec{x}) = \rho(\vec{w}^\T\vec{x} - b)$ with $(\vec{w}, b) \in \R^d \times \R$ yields
\[
    \D_t^2 \KOp \RadonOp \curly{r}(\vec{\alpha}, t) = \norm{\vec{w}}_2 \delta_{\RadonOp}((\vec{\alpha}, t) - (\tilde{\vec{w}}, \tilde{b})),
    \label{eq:multivariate-sparsified}
\]
where $\tilde{\vec{w}} = \vec{w} / \norm{\vec{w}}_2$ and $\tilde{b} = b / \norm{\vec{w}}_2$. This is analogous to the how $\D^2$ is the sparsifying transform for univariate neurons as in \cref{eq:univariate-sparsified}. The sparsifying operator is simply the second derivative in the filtered Radon domain. The key idea is that the (filtered) Radon transform allows us to extract the (univariate) activation function from the multivariate neuron and apply the univariate sparsifying transform in the $t$ variable. \Cref{fig:Radon-ReLU} is a cartoon diagram which depicts the the sparsifying transform of a ReLU neuron.

The story is now analogous to the univariate case. Indeed, by the NBT, training the NN in \cref{eq:shallow-multivariate} with weight decay is equivalent to solving the optimization problem
\[
    \min_{\vec{\theta} \in \Theta} \sum_{n=1}^N \mathcal{L}(y_n, f_\vec{\theta}(x_n)) + \lambda \sum_{k=1}^K \abs{v_k}\norm{\vec{w}_k}_2.
\]
From \cref{eq:multivariate-sparsified} we see that $\norm{\D_t^2 \KOp \RadonOp f_\vec{\theta}}_\M = \sum_{k=1}^K \abs{v_k} \norm{\vec{w}_k}_2$, and so training the NN \cref{eq:shallow-multivariate} with weight decay prefers solutions with sparse second derivatives in the filtered Radon domain. This measure of sparsity can be viewed as the second-order TV in the (filtered) Radon domain. Let $\RBV^2(\R^d)$ denote the space of functions on $\R^d$ of second-order bounded variation in the (filtered) Radon domain (i.e., the second-order TV in the (filtered) Radon domain is finite). A key result regarding $\RBV^2(\R^d)$ is the following \emph{representer theorem} for neural networks, first proven in~\cite{ParhiJMLR2021}. Under mild assumptions on the loss function, the solution set to the optimization problem
\[
    \min_{f \in \RBV^2(\R^d)} \sum_{n=1}^N \mathcal{L}(y_n, f(\vec{x}_n)) + \lambda \norm{\D_t^2 \KOp \RadonOp f}_\M
    \label{eq:RBV2-problem}
\]
is completely characterized by NNs of the form
\[
    f_\vec{\theta}(x) = \sum_{k=1}^K v_k \rho(\vec{w}_k^\T\vec{x} - b_k) + \vec{c}^\T\vec{x} + c_0,
    \label{eq:shallow-multivariate-skip}
\]
where the number of neurons is strictly less than the number of data ($K < N$) in the sense that the solution set to \cref{eq:RBV2-problem} is a closed convex set whose extreme points take the form of \cref{eq:shallow-multivariate-skip} with $K < N$
(see~\cite{BartolucciRKBS,ParhiPhD,UnserRidges} for further refinements of this result).  Common loss functions such as the squared-error satisfy the mild assumptions. The skip connection $\vec{c}^\T\vec{x} + c_0$ arises because the null space of the sparsifying transform is the space of affine functions.
Therefore, by the same argument presented in the univariate case, sufficiently wide ($K \geq N$) NNs (as in \cref{eq:shallow-multivariate-skip}) trained with weight decay to global minimizers are \emph{exactly} optimal functions in $\RBV^2(\R^d)$.

\subsection{Deep Neural Networks} \label{sec:deep-representer}
\begin{figure}[t]
    \centering
        \includegraphics[width=0.8\textwidth]{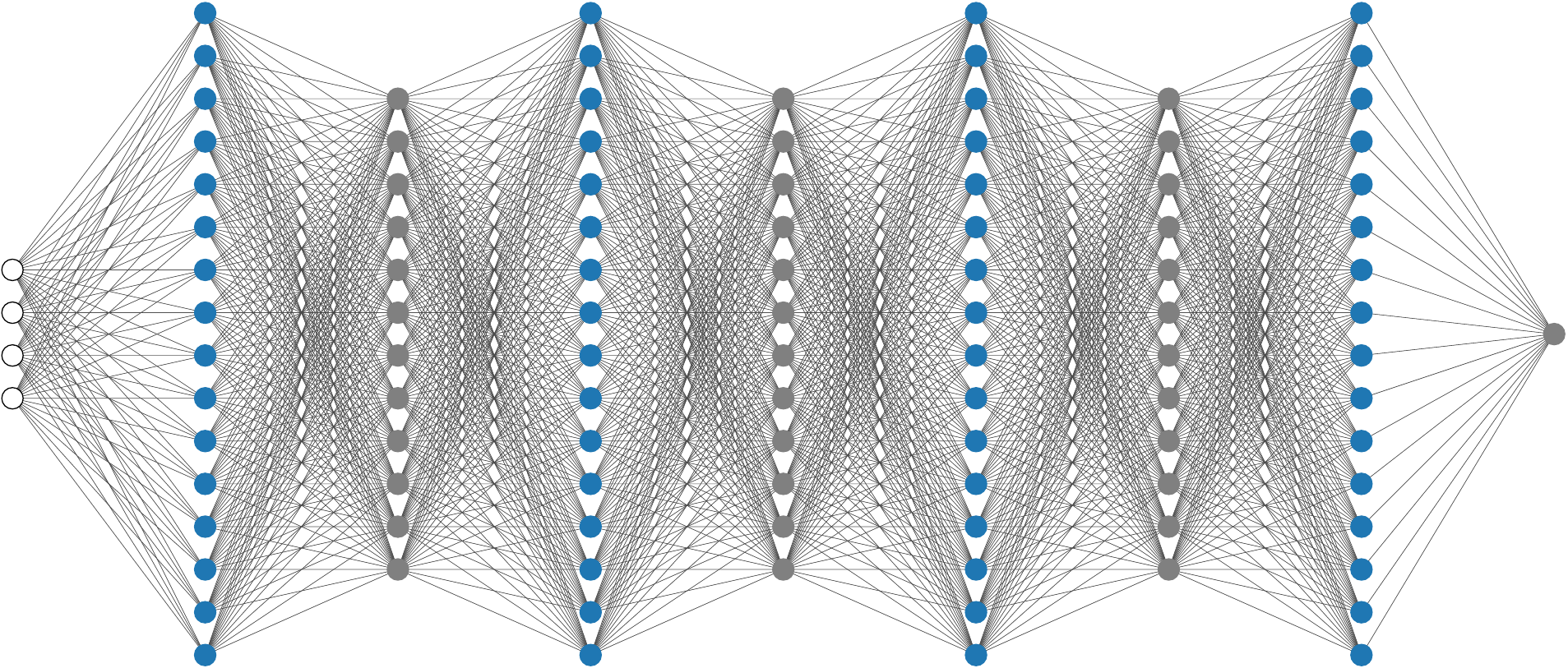}
    \caption{A feedforward DNN architecture with linear bottlenecks. The blue nodes represent ReLU neurons, the gray nodes represent linear neurons, and the white nodes depict the DNN inputs. Since the linear layers are narrower than the ReLU layers, this architecture is referred to as a DNN with linear bottlenecks.}
    \label{fig:linear-bottleneck}
\end{figure}

The machinery is straightforward to extend to the case of deep neural networks (DNNs). The key idea is to consider fitting data using  compositions of $\RBV^2$-functions. It is shown in~\cite{ParhiSIMODS2022,ShenoudaVV} that under mild assumptions on the loss function, a solution to the optimization problem
\[
    \min_{f^{(1)},\dots,f^{(L)}} \ \sum_{n=1}^N \mathcal{L}(\vec{y}_n, f^{(L)} \circ \cdots \circ f^{(1)}(\vec{x}_n)) + \lambda \sum_{\ell=1}^L \sum_{i=1}^{d_\ell} \norm{\D_t^2 \KOp \RadonOp f^{(\ell)}_i}_\M
    \label{eq:deep-problem}
\]
has the form of a DNN as in \cref{eq:ff-DNN}, where $d_\ell$ are the intermediary dimensions in the function compositions, that satisfies the following properties:
\begin{itemize}
    \item The number of layers is $L + 1$;
    \item The solution is sparse in the sense of having few active neuron connections (widths of the layers are bounded by $N^2$);
    \item The solution has skip connections in all layers;
    \item The architecture has \emph{linear bottlenecks} which forces the weight matrices to be low rank.
\end{itemize}
Such an architecture is illustrated in \cref{fig:linear-bottleneck}.
The result shows that ReLU DNNs with skip connections and linear bottlenecks trained with a variant of weight decay~\cite[Remark~4.7]{ParhiSIMODS2022} are  optimal solutions to fitting data using compositions of $\RBV^2$-functions. Linear bottlenecks may be written as factorized (low-rank) weight matrices of the form $\mat{W}^{(\ell)} = \mat{U}^{(\ell)} \mat{V}^{(\ell)}$. These bottleneck layers correspond to layers with linear activation functions ($\sigma(t) = t$). They arise naturally due to the compositions of functions that arise in \cref{eq:deep-problem}. The number of neurons in each bottleneck layers is bounded by $d_\ell$. The incorporation of linear bottlenecks of this form in DNNs have been shown to speed up learning~\cite{BaDeep} and increase the accuracy~\cite{GolubevaWider}, robustness~\cite{SanyalStable}, and computational efficiency~\cite{WangPufferfish} of DNNs.

\section{What is the Role of Neural Network Activation Functions?} \label{sec:role}
The primary focus of the article so far has been the ReLU activation function $\rho(t) = \max\curly{0, t}$. Many of the previously discussed ideas can be extended to a broad class of activation functions. The property of the ReLU exploited so far has been that it is sparsified by the second derivative operator in the sense that $\D_t^2 \rho = \delta$. Indeed, we can define a broad class of \emph{neural function spaces} akin to $\RBV^2(\R^d)$ by defining spaces characterized by different sparsifying transforms matched to an activation function. This entails replacing $\D_t^2$ in \cref{eq:sparsifying-transform} with a generic sparsifying transform $\HOp$. Table~1 (adapted from~\cite{UnserKernelNN}) provides examples of common activation functions that fall into this framework, where each sparsifying transform $\HOp$ is defined by its frequency response $\hat{H}(\omega)$. For the ReLU, we have $\HOp = \D_t^2$ and so $\hat{H}(\omega) = (\imag \omega)^2 = -\omega^2$. 

\begin{tcolorbox}[float*=t,
    width=\linewidth,
    toprule = 1mm,
    bottomrule = 0.5mm,
    leftrule = 0.5mm,
    rightrule = 0.5mm,
    arc = 0mm,
    fonttitle = \sffamily\bfseries\large,
    colframe = blue!70!black,
    title = {[Table 1] Common  Activation Functions}]
    \centering
    \begin{tabular}{rll}
        \hline \hline
        \multirow{2}{*}{Activation Function} & \multirow{2}{*}{$\sigma(t)$} & Frequency Response of \\ & & Sparsifying Transform: $\hat{H}(\omega)$ \\ \hline
        \\[-2ex]
        Rectified Linear Unit (ReLU) & $\max\curly{0, t}$ & $-\omega^2$ \\[3ex]
        Truncated Power & $\dfrac{\max\curly{0,t}^k}{k!}$, $k \in \N$ & $(\imag \omega)^{k+1}$ \\[3ex]
        Sigmoid & $\dfrac{1}{1 + e^{-t}} - \dfrac{1}{2}$ & $\dfrac{\imag}{\pi} \sinh(\pi \omega)$ \\[3ex]
        $\arctan$ & $\dfrac{\arctan(t)}{\pi}$ & $\imag \omega e^{\abs{\omega}}$ \\[3ex]
        Exponential & $\dfrac{e^{-\abs{t}}}{2}$ & $1 + \omega^2$ \\[1ex] \hline \hline
    \end{tabular}
\end{tcolorbox}

Therefore, many of the previously discussed results can thus be directly extended to a broad class of activation functions including the classical sigmoid and $\arctan$ activation functions. We remark that the sparsity-promoting effect of weight decay hinges on the homogeneity of the activation function in the DNN. While the ReLU and truncated power activation functions in Table~1 are homogeneous, the other activation functions are not. This provides evidence that one should prefer homogeneous activation functions like the ReLU in order to exploit the tight connections between weight decay and sparsity. Although the sparsity-promoting effect of weight decay does not apply to the non-homogeneous activation functions, statements akin to \cref{eq:sparsifying-transform} do hold by considering neurons with input weights constrained to be unit norm. Therefore, these sparsifying transforms uncover the innovations of finite-width NNs with unit norm input weights. Therefore, by only considering neurons with unit norm input weights, the key results which characterize the solution sets to the optimization problems akin to \cref{eq:RBV2-problem,eq:deep-problem} hold, providing insight into the kinds of functions favored by DNNs using these activation functions.

\section{Why Do Neural Networks Seemingly Break the Curse of Dimensionality?}

In 1993, Barron published his seminal paper~\cite{BarronUniversal} on the ability of NNs with sigmoid activation functions to approximate a wide variety of multivariate functions. Remarkably, he showed that NNs can approximate functions which satisfy certain decay conditions on their Fourier transforms at a rate that is \emph{completely independent} of the input-dimension of the functions. This property has led to many people heralding his work as ``breaking the curse of dimensionality''. Today, the function spaces he studied are often referred to as the \emph{spectral Barron spaces}. It turns out that this remarkable approximation property of NNs is due to sparsity.

To explain this phenomenon, we first recall a problem which ``suffers the curse of dimensionality''. A classical problem in signal processing is reconstructing a signal from its samples. Shannon's sampling theorem asserts that sampling a bandlimited signal on a regular grid at a rate faster than the Nyquist rate guarantees that the $\sinc$ interpolator perfectly reconstructs the signal. Since the $\sinc$ function and its shifts form an orthobasis for the space of bandlimited signals, the energy of the signal (squared $L^2$-norm) corresponds to the squared (discrete) $\ell^2$-norm of its samples. Multivariate versions of the sampling theorem are similar and assert that sampling multivariate bandlimited signal on a sufficiently fine regular grid guarantees perfect reconstruction with (multivariate) $\sinc$ interpolation. It is easy to see that the grid size (and therefore the number of samples) grows exponentially with with the dimension of the signal. This shows that the sampling and reconstruction of bandlimited signals suffers the curse of dimensionality. The fundamental reason for this is that the energy or ``size'' of a bandlimited signal corresponds to the $\ell^2$-norm of the signal's expansion coefficients in the $\sinc$ basis.

It turns out that there is a stark difference if we instead measure the ``size'' of a function by the more restrictive $\ell^1$-norm instead of the $\ell^2$-norm, an idea popularized by wavelets and compressed sensing. Let $\mathcal{D} = \curly{\psi}_{\psi \in \mathcal{D}}$ be a dictionary of atoms (e.g., $\sinc$ functions, wavelets, neurons, etc.). Consider the problem of approximating a multivariate function mapping $\R^d \to \R$ that admits a decomposition $f(\vec{x}) = \sum_{k = 1}^\infty v_k \psi_k(\vec{x})$, where $\psi_k \in \mathcal{D}$ and the expansion coefficients satisfy $\sum_{k=1}^\infty \abs{v_k} = \norm{\vec{v}}_{\ell^1} < \infty$. It turns out that that there exists an approximant constructed with $K$ terms from the dictionary $\mathcal{D}$ whose $L^2$-approximation error $\norm{f - f_K}_{L^2}$ decays at a rate completely independent of the input dimension $d$.

We will illustrate the argument when $\mathcal{D} = \curly{\psi_k}_{k=1}^\infty$ is an orthonormal basis (e.g., multivariate Haar wavelets). Given a function $f: \R^d \to \R$ that admits a decomposition $f(\vec{x}) = \sum_{k=1}^\infty v_k \psi_k(\vec{x})$ such that $\norm{\vec{v}}_{\ell^1} < \infty$, we can construct an approximant $f_K$ by a simple thresholding procedure that keeps the $K$ largest coefficients of $f$ and sets all other coefficients to $0$. If we let $\abs{v_{(1)}} \geq \abs{v_{(2)}} \geq \cdots$ denote the coefficients of $f$ sorted in nonincreasing magnitude, then the squared approximation error is bounded as
\[
    \norm{f - f_K}_{L^2}^2 = \norm*{\sum_{k > K} v_{(k)} \psi_{(k)}}_{L^2}^2 = \sum_{k > K} \abs{v_{(k)}}^2,
    \label{bound}
\]
where the last equality follows by exploiting the orthonormality of the $\curly{\psi_k}_{k=1}^\infty$. Finally, since the original sequence of coefficients $\vec{v} = (v_1, v_2, \ldots)$ is absolutely summable, $\abs{v_{(k)}}$ must decay strictly faster than $1/k$ for $k > K$ (since the tail of the harmonic series $\sum_{k>K} \frac{1}{k}$ diverges). Putting this together with \cref{bound}, the $L^2$-approximation error $\norm{f - f_K}_{L^2}$ must decay as $K^{-1/2}$, completely independent of the input dimension $d$. For a more precise treatment of this argument, we refer the reader to Theorem~9.10 in Mallat's \emph{Wavelet Tour of Signal Processing}~\cite{MallatBook}. These kinds of thresholding procedures, particularly with wavelet bases~\cite{DonohoWaveletShrinkage}, revolutionized signal and image processing and were the foundations of compressed sensing~\cite{CandesCS,DonohoCS}.

By a more sophisticated argument, a similar phenomenon occurs when the orthonormal basis is replaced with an essentially arbitrary dictionary of atoms. The result for general atoms is based on a probabilistic technique presented by Pisier in 1981 at the Functional Analysis Seminar at \'Ecole Polytechnique, Palaiseau, France, crediting the idea to Maurey~\cite{MaureyPisier}. An implication of Maurey's technique is that, given a function which is an $\ell^1$-combination of bounded atoms from a dictionary, there exists a $K$-term approximant that admits a dimension-free approximation rate that decays as $K^{-1/2}$. Motivated by discussions with Jones on his work on greedy approximation~\cite{JonesGreedy}, which provides a deterministic algorithm to find the approximant that admits the dimension-free rate, Barron used the technique of Maurey to prove his dimension-free approximation rates with sigmoidal NNs. This abstract approximation result is now referred to as the Maurey--Jones--Barron lemma.

In particlar, the Maurey--Jones--Barron lemma can be applied to any function space where the functions are $\ell^1$-combinations of bounded atoms. Such spaces are sometimes called \emph{variation spaces}~\cite{BachCurse,KurkovaVariation}. Recall from \cref{sec:kinds,sec:role} that the operator $\HOp \KOp \RadonOp$ sparsifies neurons of the form $\sigma(\vec{w}^\T\vec{x} - b)$, where $(\vec{w}, b) \in \cyl$ and $\sigma$ is matched to $\HOp$. This implies that the space of functions $f: \R^d \to \R$ such that $\norm{\HOp \KOp \RadonOp f}_\M < \infty$ can be viewed as a variation space, where the dictionary corresponds to the neurons $\curly{\sigma(\vec{w}^\T\vec{x} - b)}_{(\vec{w}, b) \in \cyl}$. Therefore, given $f: \R^d \to \R$ such that $\norm{\HOp \KOp \RadonOp f}_\M < \infty$, there exists a $K$-term approximant $f_K$ that takes the form of a shallow NN with $K$ neurons such that the $L^2$-approximation error decays as $K^{-1/2}$. 
These techniques have been studied and extended in great detail~\cite{SiegelSharp} and have been extended to the setting of deep NNs~\cite{WojtowytschDeepBanach} by considering compositional function spaces akin to the compositional space introduced in \cref{sec:deep-representer}.

Combining these dimension-free approximation rates with the sparsity-promoting effect of weight decay regularization for ReLU NNs has a striking effect on the learning problem. Suppose that we train a shallow ReLU NN with weight decay on data generated from the noisy samples $y_n = f(\vec{x}_n) + \varepsilon_n$, $n = 1, \ldots, N$, of $f \in \RBV^2(\R^d)$, where $\vec{x}_n$ are i.i.d.\ uniform random variables on some bounded domain $\Omega \subset \R^d$ and $\varepsilon_n$ are i.i.d.\ Gaussian random variables. Let $f_N$ denote this trained NN. Then, it has been shown~\cite{ParhiTIT2023} that the mean integrated squared error (MISE) $\mathbb{E}\norm{f - f_N}_{L^2(\Omega)}^2$ decays at a rate of $N^{-1/2}$, independent of the input dimension $d$. Moreover, this result also shows that the generalization error of the trained NN on a new example $\vec{x}$ generated uniformly at random on $\Omega$ is also immune to the curse of dimensionality. Furthermore, these ideas have been studied in the context of deep NNs~\cite{SchmidtDeepReLU}, proving dimension-free MISE rates for estimating H\"older functions (that exhibit low-dimensional structure) with ReLU DNNs.

\subsection{Mixed Variation and Low-Dimensional Structure}
\label{mixed}
\begin{figure}[t]
    \centering
    \begin{minipage}[b]{0.45\linewidth}
        \centering
        \centerline{\includegraphics[width=\textwidth]{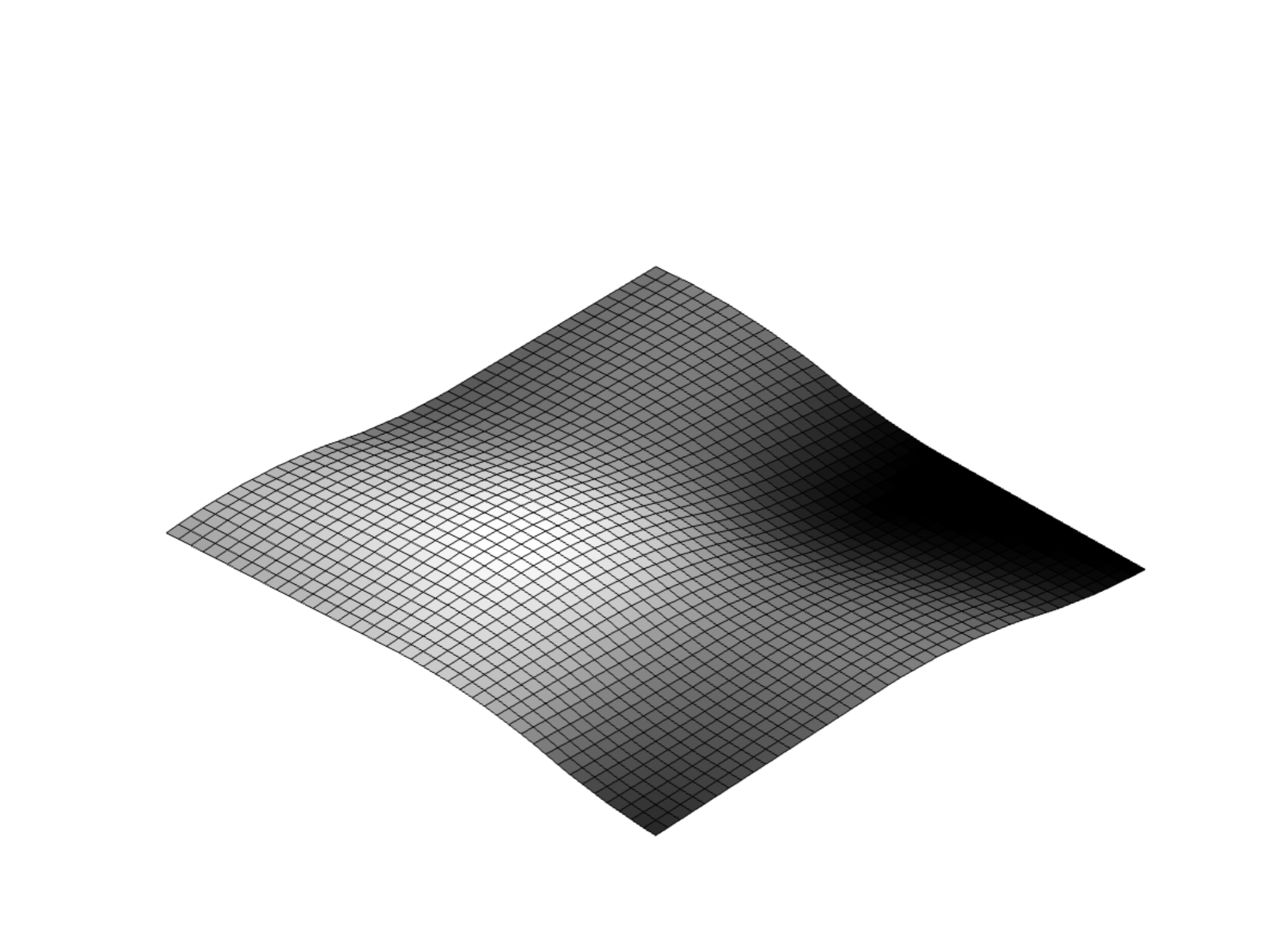}}
        (a) Weak variation in multiple directions.
    \end{minipage}
    \begin{minipage}[b]{0.45\linewidth}
        \centering
        \centerline{\includegraphics[width=\textwidth]{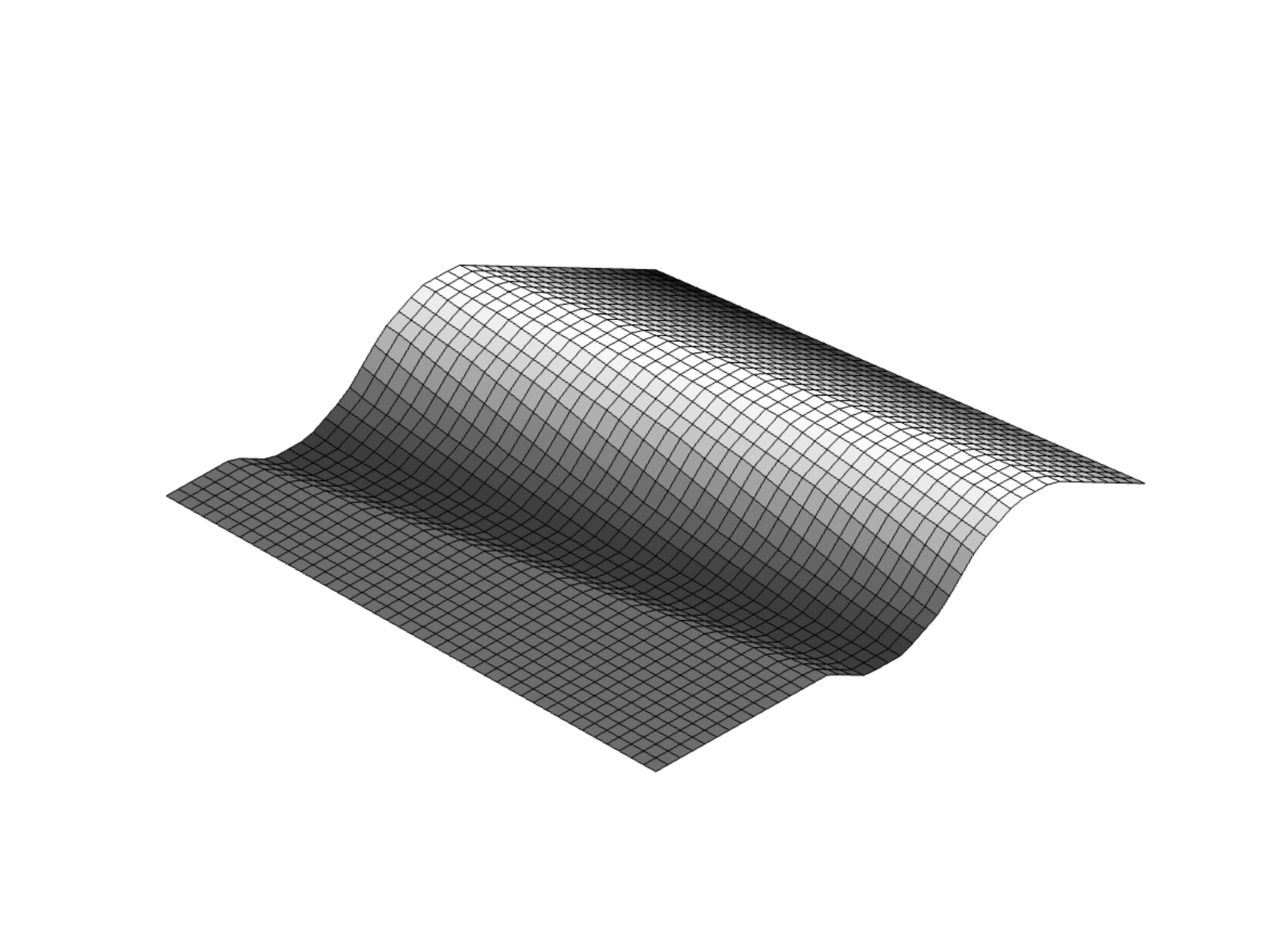}}
        (b) Strong variation in one direction.
    \end{minipage}
    \begin{minipage}[b]{0.45\linewidth}
        \centering
        \centerline{\includegraphics[width=\textwidth]{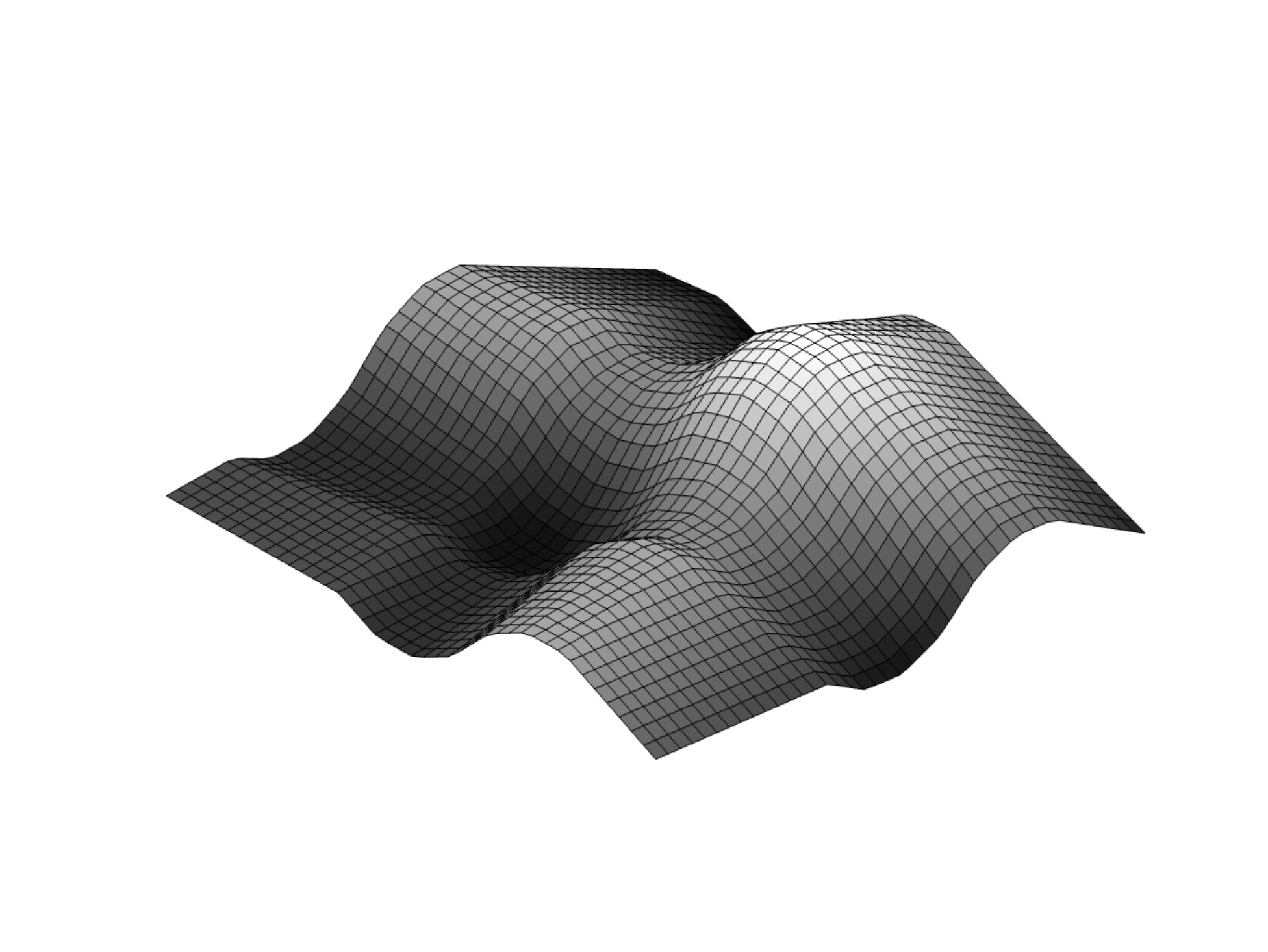}}
        (c) Strong variation in multiple directions.
    \end{minipage}
    \caption{Examples of functions exhibiting different kinds of variation.}
    \label{fig:mixed-variation}
\end{figure}

The national meeting of the American Mathematical Society in 2000 was held to discuss the mathematical challenges of the 21st Century. Here, Donoho gave a lecture titled \emph{High-Dimensional Data Analysis: The Curses and Blessings of Dimensionality}~\cite{DonohoCurse}. In this lecture, he coined the term ``mixed variation'' to refer to the kinds of functions that lie in variation spaces, citing the spectral Barron spaces as an example. The variation spaces are different from classical multivariate function spaces in that they favor functions that have weak variation in multiple directions (very smooth functions) as well as functions that have very strong variation in one or a few directions (very rough functions). These spaces also \emph{disfavor} functions with strong variation in multiple directions. It is this fact that makes them quite ``small'' compared to classical multivariate function spaces, giving rise to their dimension free approximation and MISE rates. Examples of functions with different kinds of variation are illustrated in \cref{fig:mixed-variation}.
The prototypical examples of functions that lie in mixed variation spaces can be thought of as superpositions of few neurons with different directions or superpositions of many neurons (even continuously many) in only a few directions.

In order to interpret the idea of mixed variation in the context of modern data analysis and DL, we turn our attention to \cref{fig:mixed-variation}(b). In this figure, the function has strong variation, but only in a single direction. In other words, this function has \emph{low-dimensional structure}.
It has been observed by a number of authors that DNNs are able to automatically adapt to the low-dimensional structure that often arises in natural data.
This is possible because the input weights can be trained to adjust orientation of each neuron. The dimension-independent approximation rate quantifies the power of this tunability. This explains why DNNs are good at learning functions with low-dimensional structure. In particular, the function in \cref{fig:mixed-variation}(c) has strong variation in all directions, so no method can overcome the curse of dimensionality in this sort of situation. On the other hand, in \cref{fig:mixed-variation}(a) the function has weak varation in all directions and \cref{fig:mixed-variation}(b) has strong variation only in one direction, so these are functions for which neural networks will overcome the curse. For \cref{fig:mixed-variation}(b) the the sparsity-promoting effect of weight decay promotes DNN solutions with neurons oriented in the direction of variation (i.e., it automatically learns the low-dimensional structure).

\section{Takeaway Messages and Future Research Directions}

In this article we presented a mathematical framework to understand DNNs from first principles, through the lens of sparsity and sparse regularization. Using familiar mathematical tools from signal processing, we provided an explanation for the sparsity-promoting effect of the common regularization scheme of weight decay in neural network training, the use of skip connections and low-rank weight matrices in network architectures, and why neural networks seemingly break the curse of dimensionality. This framework provides the mathematical setting for many future research directions. 

The framework suggests the possibility of new neural training algorithms. The equivalence of solutions using weight decay regularization and the regularization in \cref{eq:ff-DNN-sum-of-paths} leads to the use of proximal gradient methods akin to iterative soft-thresholding algorithms to train DNNs. This avenue has already begun to be explored. The preliminary results in~\cite{YangWeightDecayProx} have shown that proximal gradient training algorithms for DNNs perform as good as and often better (particularly when labels are corrupted) than standard gradient-based training with weight decay, while simultaneously producing sparser networks.

There has also been a large body of work dating back to 1989~\cite{LeCunBrainDamage} on NN pruning has shown empirically that large NNs can be compressed or sparsified to a fraction of their size while still maintaining their predictive performance. The connection between weight decay and sparsity-promoting regularizers like in \cref{eq:ff-DNN-sum-of-paths} suggests new approaches to pruning. For example, one could apply proximal gradient algorithms to derive sparse approximations to large pre-trained neural networks~\cite{ShenoudaVV}. There are many open questions in this direction, both experimental and theoretical, including applying these algorithms to other DNN architectures and deriving convergence results for these algorithms.  

The framework in this paper also shows that trained ReLU DNNs are compositions of $\RBV^2$-functions. As we have seen in this article, at this point in time, we have a clear and nearly complete understanding of $\RBV^2(\R^d)$. In particular, the $\RBV^2$-space favors functions that are smooth in most or all directions, which explains why neural networks seemingly break the curse of dimensionality. Less is clear and understood about the compositions of $\RBV^2$-functions (which characterize DNNs). Better understanding of compositional function spaces could provide new insights into the benefits of depth in neural networks. This in turn could lead to new guidelines for designing NN architectures and training algorithms.

\section*{Acknowledgment}
The authors would like to thank
Rich Baraniuk,
Misha Belkin,
Çağatay Candan,
Ron DeVore,
Kangwook Lee,
Greg Ongie,
Dimitris Papailiopoulos,
Tomaso Poggio,
Lorenzo Rosasco,
Joe Shenouda,
Jonathan Siegel,
Ryan Tibshirani,
Michael Unser,
Becca Willett,
Stephen Wright,
Liu Yang,
and Jifan Zhang
for many insightful discussions on the topics presented in this article.

RP was supported by in part by the U.S. National Science Foundation (NSF)  Graduate  Research  Fellowship  Program  under  grant  DGE-1747503 and the European Research Council (ERC Project FunLearn) under Grant 101020573. RN was supported in part by the NSF grants DMS-2134140 and DMS-2023239, the ONR MURI grant N00014-20-1-2787, and the AFOSR/AFRL grant FA9550-18-1-0166, as well as the Keith and Jane Nosbusch Professorship.

\bibliographystyle{IEEEtranS}
\bibliography{ref}

\begin{IEEEbiographynophoto}{Rahul Parhi (rahul.parhi@epfl.ch)}
received the B.S. degree in mathematics and the B.S. degree in computer science from the University of Minnesota--Twin Cities in 2018 and the M.S. and Ph.D. degrees in electrical engineering from the University of Wisconsin--Madison in 2019 and 2022, respectively. During his Ph.D., he was supported by an NSF Graduate Research Fellowship. He is currently a Post-Doctoral Researcher with the Biomedical Imaging Group at the \'Ecole Polytechnique F\'ed\'erale de Lausanne. He is primarily interested in applications of functional and harmonic analysis to problems in signal processing and data science. He is a member of the IEEE.
\end{IEEEbiographynophoto}

\begin{IEEEbiographynophoto}{Robert D. Nowak (rdnowak@wisc.edu)} holds a Ph.D. in electrical engineering from the University of Wisconsin--Madison and is currently the Grace Wahba Professor of Data Science and Keith and Jane Morgan Nosbusch Professor in Electrical and Computer Engineering at the same institution. With research focusing on signal processing, machine learning, optimization, and statistics, Nowak's work on sparse signal recovery and compressed sensing has won several awards. He currently serves as a Section Editor for the SIAM Journal on Mathematics of Data Science and a Senior Editor for the IEEE Journal on Selected Areas in Information Theory, and is an IEEE Fellow.
\end{IEEEbiographynophoto}

\end{document}